\documentclass[conference]{IEEEtran}
\IEEEoverridecommandlockouts
\usepackage{cite}
\usepackage{amsmath,amssymb,amsfonts}
\usepackage{algorithmic}
\usepackage{graphicx}
\usepackage{textcomp}
\usepackage{xcolor}

\usepackage[numbers]{natbib}
\usepackage{hyperref}

\newif\ifappendix
\appendixtrue 

\newif\ifproof
\prooffalse 
\prooftrue 

\usepackage{microtype}
\usepackage{graphicx}
\usepackage{subfigure}
\usepackage{multirow} 
\usepackage{multicol} 
\usepackage{mathtools}
\usepackage{xspace}
\usepackage{amsmath}
\usepackage{amsfonts}
\usepackage{hyperref}
\usepackage{color}
\usepackage{xcolor}
\usepackage{listings}
\usepackage{lstbayes}

\newif\ifappendix
\appendixfalse 
\appendixtrue 

\newif\ifproof
\prooffalse 
\prooftrue 

\newif\ifstan
\stanfalse

\def \fvec {\text{\boldmath$f$}}    
\def \wvec {\text{\boldmath$w$}}

\def \xvec {\text{\boldmath$x$}}

\def \zvec {\text{\boldmath$z$}} 
\def \mZ {\text{$\mathbf Z$}}

\def \thetavec {\text{\boldmath$\theta$}}

\def \betavec {\text{\boldmath$\beta$}}
\def \alphavec {\text{\boldmath$\alpha$}}

\def \varthetavec {\text{\boldmath$\vartheta$}}
\def \lambdavec {\text{\boldmath$\lambda$}}

\DeclareMathOperator{\softplus}{softplus}
\DeclareMathOperator{\Be}{Be}
\DeclareMathOperator{\Ber}{Ber}
\DeclareMathOperator{\ELBO}{ELBO}
\DeclareMathOperator{\KL}{KL}

\DeclarePairedDelimiterX{\infdivx}[2]{(}{)}{%
  #1\;\delimsize\|\;#2%
}
\newcommand{\dkl}[2]{\ensuremath{\KL\infdivx{#1}{#2}}\xspace}

\begin{document}

\title{Bernstein Flows for Flexible Posteriors in Variational Bayes}

\author{
\IEEEauthorblockN{Oliver D{\"u}rr \IEEEauthorrefmark{1}} 
\IEEEauthorblockA{IOS, Konstanz University of Applied Sciences\\
Email: oliver.duerr@htwg-konstanz.de}
\and
\IEEEauthorblockN{Stephan Hörling}
\IEEEauthorblockA{IOS, Konstanz University of Applied Sciences \\
Email: Stefanhoertling@web.de}
\and
\IEEEauthorblockN{Daniel Dold }
\IEEEauthorblockA{IOS, Konstanz University of Applied Sciences \\
Email: ddold@htwg-konstanz.de}
\and
\IEEEauthorblockN{Ivonne Kovylov}
\IEEEauthorblockA{IOS, Konstanz University of Applied Sciences \\
Email: ivonne97ko@gmail.com}
\and
\IEEEauthorblockN{Beate Sick \IEEEauthorrefmark{1}}
\IEEEauthorblockA{EBPI, University of Zurich \& \\
IDP, Zurich University of Applied Sciences\\
Email: beate.sick@uzh.ch, sick@zhaw.ch}

\IEEEauthorblockA{\IEEEauthorrefmark{1} corresponding authors, contributed equally}
}

\title{Bernstein Flows for Flexible Posteriors in Variational Bayes}

\maketitle

\begin{abstract}
Variational inference (VI) is a technique to approximate difficult to compute posteriors by optimization. In contrast to MCMC, VI scales to many observations. In the case of complex posteriors, however, state-of-the-art VI approaches often yield unsatisfactory posterior approximations. This paper presents Bernstein flow variational inference (BF-VI), a robust and easy-to-use method, flexible enough to approximate complex multivariate posteriors. BF-VI  combines ideas from normalizing flows and Bernstein polynomial-based transformation models.  In benchmark experiments, we compare BF-VI solutions with exact posteriors, MCMC solutions, and state-of-the-art VI methods including normalizing flow based VI.  We show for low-dimensional models that BF-VI accurately approximates the true posterior; in higher-dimensional models, BF-VI outperforms other VI methods. Further, we develop with BF-VI a Bayesian model for the semi-structured Melanoma challenge data, combining a CNN model part for image data with an interpretable model part for tabular data, and demonstrate for the first time how the use of VI in semi-structured models.
\end{abstract}

\section{Introduction}
\label{sec:intro}
Uncertainty quantification is essential, especially if model predictions are used to support high-stakes decision-making. Quantifying uncertainty in statistical or machine learning models is often achieved by Bayesian approaches, where posterior distributions represent the uncertainty of the estimated model parameters. Determining the exact posterior distributions is often impossible when the posterior takes a complex shape and the model has many parameters. This is especially true for complex models such as Bayesian neural networks (NNs) or semi-structured models that combine an interpretable model part with deep NNs. Variational inference (VI) is a commonly used approach to approximate complex distributions through optimization \cite{jordan1999, bleiVI}. In VI, 
the complex posterior is approximated by a variational distribution by minimizing a divergence measure between the variational and the true posterior distribution.
VI is currently a very active research field tackling different challenges, which can be categorized into the following groups: (1) constructing variational distributions that are flexible enough to match the true posterior distribution,
(2) defining optimal variational objective for tuning the variational distribution, which boils down to finding the most suited divergence measure quantifying the difference between a variational distribution and posterior, and (3) developing robust and accurate stochastic optimization frameworks for the variational objective \cite{dhaka2020robust, blei2016variational, welandawe2022robust}. 
Here, we focus on challenge (1) and propose a method to construct a variational distribution that is flexible enough to accurately and robustly approximate complex multidimensional posteriors. 

To avoid model-specific calculations, we design our method as a Black Box VI (BBVI) approach \cite{ranganath2014black}. 
In BBVI, the approximative posterior is determined by stochastic gradient descent. The user simply defines the Bayesian model by specifying the likelihood and the prior, after which all subsequent calculations are carried out automatically.  
Due to its simplicity, BBVI is implemented in many packages for Bayesian modeling, like Stan~\cite{carpenter2017stan} and Pyro~\cite{bingham2019pyro} as an alternative to MCMC. Given BBVI's scalability to large datasets and its widespread applicability, it has emerged as the preferred technique in the field of machine learning~\cite{welandawe2022robust}.

Our approach uses transformation models (TMs) to construct complex posteriors.
Transformation models (TMs) have been introduced for fitting potentially complex outcome distributions for probabilistic regression models \cite{hothorn2014}. Since then, they have been mainly used to model different outcome types, such as ordinal \cite{kook_herzog2020, buri2020b}, count \cite{Siegfried_Hothorn_2020}, continuous \cite{lohse2017continuous}, or time-to-event outcomes \cite{campanella2022deep} based on tabular predictors. Moreover, TMs have been used to model multidimensional distributions \cite{klein2019multivariate}. 
Neural networks can be used to extend TMs to model outcomes for unstructured predictors (e.g., images or text) or a combination of tabular and unstructured predictors \citep{sick2021, baumann2020deep, kook_herzog2020, rugamer2021timeseries}. 

The basic idea of TMs is to learn a flexible and monotone transformation function that transforms between a simple latent distribution and a potentially complex conditional outcome distribution.  
In TMs, the transformation function is parameterized as an expansion of basis functions. In the case of continuous target distributions, most often, Bernstein polynomials~\cite{bernvstein1912demonstration} are used because they can easily be constrained to be strictly monotone, and their flexibility can be tuned via the order $M$. A large order $M$ ensures an accurate approximation of the distribution, which is robust against a further increase of $M$ \cite{hothorn2018, ramasinghe2021robust}; this is also demonstrated in our experiments for the BBVI setting. 

Independently of TMs, normalizing flows (NFs) have been developed in the deep learning community. NFs and TMs rely on the same idea, but NFs usually construct the transformation by chaining many simple functions, while TMs construct one rather complex transformation function. 
In NFs, each simple function, such as shifting and scaling, incrementally adds to the complexity of the final transformation. Among the prominent NF implementations are RealNVP~\cite{dinh2016density} and Masked Autoregressive Flow (MAF)~\cite{papamakarios2017masked}. RealNVP stands out for its efficient, invertible transformations facilitated by a specialized neural network architecture. 
Its key advantage lies in the efficient computation of the Jacobian matrix's determinant, essential for direct density estimation in the change of variable function (see \ref{eq:vi_change}). 
This efficiency is achieved by iteratively splitting the components of the data into two parts. In each step, the first part of the components is used to train a neural network computing the scale and shift parameters of the transformation. This transformation is then applied to the other components, while the first part remains unaltered. This procedure is repeated multiple times with different partitioning, leading to a triangular Jacobian, thus enabling efficient and invertible transformations.
In contrast, MAF adopts a fundamentally different approach to construct transformations~\cite{papamakarios2017masked}. It utilizes a sequential (autoregressive) framework, facilitated by neural networks. In MAF each output component relies exclusively on its preceding components, a concept often referred to as causality in this context. This design also leads to a triagonal Jacobian matrix and thus a fast computation of the change of variable equation.
The MAF ensures that the \(n\)th output of NN is solely dependent on the first \(n-1\) inputs, yielding an autoregressive model. 
However, some NF approaches use a single flexible transformation, such as sum-of-squares polynomials \cite{jaini2019sum} or splines \cite{durkan2019neural}. Recently, also Bernstein-based polynomials have been used for modeling unconditional multivariate density distributions \cite{ramasinghe2021robust}. 

%
%

NFs were initially introduced for variational inference to approximate potentially complex distributions of latent variables in models such as variational autoencoders \cite{rezende2015, van2018sylvester}. In the past, oftenmembers from simple distribution families have been used to approximate the posterior in BBVI. In the "Bayes by Backprop" method, Blundell et al. used independent Gaussians to approximate the posterior of the weights in a Bayesian Neural Network (BNN). They determined the parameters of these Gausians using BBVI~\cite{blundell2015}. This approach was made more flexible by using a multivariate Gaussians~\cite{louizos2017multiplicative} as variational distribution. While it is clear that TMs or NFs have the potential to construct flexible variational distributions, the first attempts to use NF-based BBVI were proposed only recently~\cite{agrawal2020}. These NF-based BBVI approaches compare favorably against existing BBVI methods but require a complex training scheme, and sometimes exhibit pathological behavior~\cite{dhaka2021challenges}. 
%

Here, we introduce Bernstein flow variational inference (BF-VI), which, for the first time, uses TMs in BBVI. We use TMs based on Bernstein polynomials to construct a variational distribution that closely approximates a potentially complex posterior in Bayesian models. The proposed method is computationally efficient and applicable to typical statistical models. The proposed method yields superior results in our experiments compared to existing NF approaches  ~\cite{dhaka2021challenges}. 
Using BF-VI we further demonstrate, for the first time, how VI can be used to fit Bayesian semi-structured models where interpretable statistical model parts (based on tabular data) and deep NN model parts (based on images) are jointly fitted.  
We define our method in section \ref{sec:tm} for one-dimensional examples and generalize it to Bayesian models with multivariate posteriors in section \ref{sec:multivar}. 
In section \ref{sec:results}, we benchmark our BF-VI approach against exact Bayesian models, MCMC-Simulations, Gaussian-VI, and NF-based BBVI, showing accurate posterior approximations in low dimensions and superior approximations in higher dimensions when compared to NF-based BBVI and summarize in section \ref{sec:outlook}.

\section{Bernstein Flow Variational Inference}
\label{sec:methods}
In the following, we describe the Bernstein Flow-VI (BF-VI) approach, which we propose for accurately and robustly approximating potentially complex posteriors in Bayesian models. 
The main idea is to enable the VI procedure to approximate the joined posterior of the $p$ model parameters by a flexible variational distribution. This is done by modeling the transformation function from a predefined simple latent distribution to a potentially complex variational distribution. The number of parameters $p$ in the Bayesian model determines the dimension of both the latent and the variational distribution.

We first explain BF-VI for Bayesian models with a single parameter and hence a one-dimensional posterior and then generalize to  models with multivariate posteriors. 
The code is publicly available on GitHub\footnote{
\url{https://github.com/tensorchiefs/bfvi_paper}
}. 
\subsection{One-Dimensional Bernstein Flows}
\label{sec:tm}
\begin{figure}[!h]
    \centering
    \includegraphics[width=0.75\columnwidth]
    {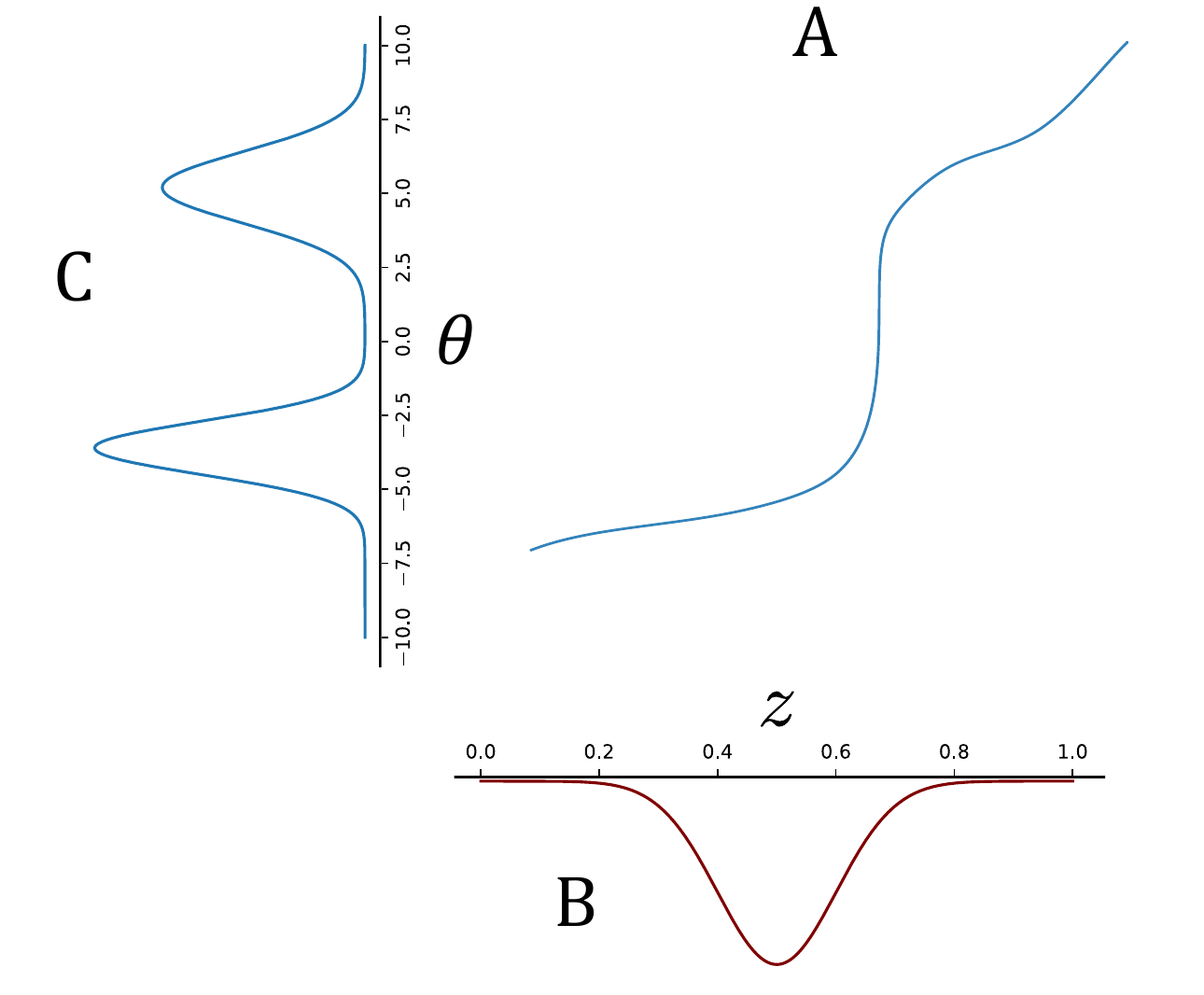}
    \caption{Overview of the transformation model. A: shows the bijective transformation function $g: Z  \rightarrow \theta$ (or its approximation $f_\text{BP}$) mapping between B: a predefined latent density $f_Z$ and C: a potentially complex posterior (or its variational distribution).}
    \label{fig:flow}
\end{figure}

BF-VI approximates the bijective transformation  function $g: Z  \rightarrow \theta$ between a latent variable $Z \in \mathbb{R}$ with predefined  
distribution  $F_Z : \mathbb{R} \to [0, 1]$ with log-concave and 
continuous density $f_Z$, 
and the model parameter $\theta  \in \mathbb{R}$ with a potentially complex distribution $F_\theta: \mathbb{R} \to [0, 1]$ so that $F_Z(z)=F_\theta(g(z))$. Fig.~\ref{fig:flow} visualizes this transformation on the scale of the densities, where $f_Z(z)=f_\theta(g(z))  \mid \frac{\partial{g(z)}}{\partial{z}} \mid$ according to the change-of-variable formula.

Hothorn et al. \cite{hothorn2018} give theoretical guarantees for the existence and uniqueness of $g=F_\theta^{-1} \circ F_Z$.  However, $g$ cannot be computed directly if $F_\theta$ is not known (in our application $F_\theta$ is the unknown distribution of the posterior).
The core of BF-VI is to approximate $g$, shown in Fig.~\ref{fig:flow}, by $f_\text{BP}$ Bernstein polynomials (BP)\footnote{ 
Some authors make a distinction between Bernstein polynomials in which $\vartheta_i$ is fixed by the values of the function to be approximated and call expressions like expressions like in (\ref{eq:mlt}), where $\vartheta_i$ is a fitting parameter, polynomials of Bernstein type.}
as
\begin{equation}
\small
\label{eq:mlt}
    f_\text{BP}(z) = \sum_{i=0}^M {\Be_i(z) \frac{\vartheta_i}{M+1}}
\end{equation}
with $\Be_i(z) = \Be_{(i+1,M-i+1)}(z)$ being the density of a Beta distribution with parameters $i+1$ and $M-i+1$.  To preserve the bijectivity of $g$, we use in BF-VI w.l.o.g. a strict monotone increasing BP to approximate $g$. With the approximation of the transformation function $g$ by $f_\text{BP}$, it holds that $f_Z(z)$ can be approximated by $f_\theta(f_\text{BP}(z)) \mid \frac{\partial{f_\text{BP}(z)} }{\partial{z}} \mid$.

Using a BP for approximating $g$ gives the following theoretical guarantees \cite{farouki2012bernstein} 1) With increasing order $M$ of the BP, the approximation $f_\text{BP}$ to $g$ gets arbitrarily close (the BP have been introduced for this very purpose in the constructive proof of the Weierstrass theorem by~\cite{bernvstein1912demonstration}); 2) the required strict monotonicity of the approximation $f_\text{BP}$ can be easily achieved by constraining the coefficients $\vartheta_i$ of the BP to be increasing; 3) BPs are robust versus perturbations of the coefficients $\vartheta_i$; 4) the approximation error decreases with $1/M$ (Voronovskaya Theorem). See \cite{bernvstein1912demonstration, farouki2012bernstein} for more detailed discussions of the beneficial properties of BPs in general and \cite{hothorn2018,  ramasinghe2021robust} for transformation models.

While the output of $f_\text{BP}(z)$ is unrestricted, a BP requires a input $z$ within $[0,1]$. We experimented with several approaches to ensure the restriction $z \in [0,1]$ resulting in slightly different behavior during the training (see Appendix \ref{appendix:trafo_design}). Based on these experiments, we decided to obtain $z \in [0,1]$ by sampling values  from a standard normal distribution, $z''\sim N(0,1)$, then apply the affine transformation  $l(z'') = \alpha \cdot z'' + \beta$, followed by a sigmoid $\sigma(z')=1/(1+e^{-z'})$. Altogether,  we approximate the transformation $g$ by $f: Z  \rightarrow \theta$ via $f=f_\text{BP} \circ \sigma \circ l$, which we call Bernstein flow.

 To allow the application of unconstrained stochastic gradient descent optimization, which is typically used in the deep learning domain, we enforce the strict monotonicity of the flow $f$ as follows: We optimize unrestricted parameters of $f$, i.e., $\vartheta_0', \ldots \vartheta_{M}'$, $\alpha', \beta'$,
 and apply the following transformations to determine the parameters of the bijective flow: $\vartheta_0=\vartheta_0'$, and $\vartheta_i = \vartheta_{i-1} + \softplus(\vartheta_{i}')$ for $i=1,\dots,M$ for getting a strictly increasing BP and $\alpha = \softplus(\alpha')$, $\beta = \beta'$ for getting an increasing affine transformation.
 

 \ifappendix
 In the Appendix \ref{sec:proof}, we show that the resulting variational distribution is a tight approximation to the posterior in the sense that the KL divergence between $q_{\lambda}(\theta)$ and $p(\theta\mid D)$ decreases with the order of the BP via $1/M$.
\fi
 
\subsection{Multivariate Generalization  \label{sec:multivar}}
In the case of a Bayesian model with $p$ parameters, $\theta_1, \theta_2, \ldots, \theta_p$, the Bernstein flow bijectively maps a $p$-dimensional $\mZ'$ to a $p$-dimensional $\thetavec{}$. 
We realize this flow by choosing $p$ independent standard normal Gaussians as simple latent distribution for the p-dimensional $\mZ'$ and apply on each component an affine transformation followed by a sigmoid function to achieve a [0,1] restricted $\mZ$. The possible complex dependencies in $\thetavec{}$ are modeled in the multivariate generalization $\fvec_\text{BP}$ of the one-dimensional Bernstein Polynomial (see eq.~\ref{eq:multi_bp} for the definition of the $j$-th component of $\fvec_\text{BP}$). 
\begin{equation}
\small
\label{eq:multi_bp}
       \theta_j  =f_{{\text{BP}}_{j}}(z_{1:
    j})=\frac{1}{M+1} \sum_{i=0}^M \vartheta_i^j(z_1, \ldots, z_{j-1}) \Be_i (z_j)
\end{equation}
To achieve an efficient computation, we use a triangular map for constructing coefficients ${\vartheta_i^j}$ $j = 2,\ldots p\;, i= 0,\cdots, M$ from $\mZ$. This ensures that the $j$-th BP determining $\theta_j$ only depends on the first $j$-1 components of $\mZ$ (see eq.~\ref{eq:multi_bp}). It is known that bijective triangular maps with sufficient flexibility can map a simple $p$-dimensional distribution into arbitrary complex $p$-dimensional target distributions \cite{bogachev2005triangular}.  
We use a masked autoregressive flow (MAF)~\cite{papamakarios2017masked} to map $\mZ$ to the BP coefficients  ${\vartheta_i^j}$ $j = 2,\ldots p\;, i= 0,\cdots, M$ from $\mZ$. 
The MAF architecture ensures that 
%
that ${\vartheta_i^j}$ depend only on those components of the latent variables $z_{j'}$ with $j' \le j$ (as required in eq.~\ref{eq:multi_bp}).
Note that the first coefficients in all BPs $\varthetavec^1$ do not depend on $z$ and are therefore not modeled via the MAF.
Therefore, 
the Jacobian $\nabla{\fvec_{\text{BP}}}$ w.r.t.~$\zvec$ is a triangular matrix, and hence  $\det\nabla{\fvec_{\text{BP}}}$ is given by the product of the diagonal elements of the Jacobian allowing for efficient computation of the resulting $p$-dimensional variational distribution $q_\lambdavec(\thetavec)$ via the multivariate version of the change of variable formula (see eq.~\ref{eq:vi_change}). 
The flexibility of such a $p$-dimensional bijective Bernstein flow is only limited by the order $M$ of the Bernstein polynomial and the complexity of the MAF. 
In our experiments, we use an MAF with two hidden layers, each with 10 neurons. The weights $\wvec$ of the MAF are part of the variational parameters for  $\fvec_{\text{BP}}$.  In total, we have $\lambdavec=(\varthetavec^1,\wvec, \alphavec{}, \betavec{})$ variational parameters.

\subsection{Variational Inference Procedure} \label{sec:vi}
In VI the variational parameters $\lambdavec$
are tuned such that the resulting variational distribution $q_\lambdavec(\thetavec)$ is as close to the posterior $p(\thetavec\mid D)$ as possible. Here, we do this by minimizing the KL-divergence between the variational distribution and the (unknown) posterior:
\begin{multline}
\small
\dkl{q_\lambdavec(\thetavec)}{p(\thetavec\mid D)} = 
\int~q_\lambdavec(\thetavec)\log\left(\frac{q_\lambdavec(\thetavec)}{p(\thetavec \mid D)}\right)d\thetavec \\=
\log(p(D))  
- \underbrace{
\left( \mathbb{E}_{\thetavec \sim q_\lambdavec} (\log(p(D \mid \thetavec))) - \dkl{q_\lambdavec(\thetavec)}{p(\thetavec)}  \right)
}_{\ELBO(\lambdavec)}
\label{eq:vi1}    
\end{multline}
The KL-divergence is commonly used in VI, and a recent study showed that it is easier to train than other divergences and applicable to higher-dimensional distributions \cite{dhaka2021challenges}. 
 

Instead of minimizing (\ref{eq:vi1}) usually only the 
evidence lower bound (ELBO) is maximized \cite{blundell2015weight}  which consists of the expected value of the log-likelihood, $\mathbb{E}_{\thetavec \sim q_\lambdavec} (\log(p(D\mid\thetavec)))$, minus the KL-divergence between the variational distribution $q_\lambdavec(\thetavec)$ and the  prior $p(\thetavec)$. Note that the ELBO does not explicitly contain the unknown posterior. In practice, we minimize the negative ELBO using stochastic gradient descent facilitated by automatic differentiation. For consistency with \cite{dhaka2021challenges}, we use TensorFlow's RMSprop in all our experiments, configured with the default settings. We follow the BBVI approach and approximate the expected log-likelihood by averaging over $S$ samples $\thetavec_s \sim q_\lambdavec(\thetavec)$ via 
\begin{equation}
\small
\mathbb{E}_{\thetavec \sim q_\lambdavec} (\log(p(D_i\mid\thetavec))) \approx \frac{1}{S} \sum_{s,i} \log\left(p(D_i\mid\thetavec_s)\right).
\label{eq:vi}    
\end{equation}
Hereby, we also assume the usual independence of the $i=1,\ldots N$ training data points $D_i$. 
To get these samples $\thetavec_s$, we use $S$ samples $\zvec_s'$ from the latent distribution, apply the transformation
$f=f_\text{BP} \circ \sigma \circ l$,
and then compute the corresponding parameter samples via $\thetavec_s = \fvec(\zvec_s)$. We use the same samples $\thetavec_s \sim q_\lambdavec(\thetavec)$ to approximate the Kullback-Leibler divergence between the variational distribution $q_\lambdavec(\thetavec)$ and the prior $p(\thetavec)$ via:
\begin{equation}
\small
\dkl{q_\lambdavec(\thetavec)}{p(\thetavec)} \approx 
 \frac{1}{S} \sum_{s} \log\left(\frac{q_\lambdavec(\thetavec_s)}{p(\thetavec_s)}\right)
\label{eq:vi_sample_KL}    
\end{equation}
where the probability density $q_\lambdavec(\thetavec_s)$ can be calculated, from the samples $\thetavec_s$ using the change of variable function as:

\begin{equation}
\small
q_\lambdavec(\thetavec_s) = p(\zvec'_s) \cdot \mid 
\det\nabla_{\zvec'}{\fvec_\text{BP}(\sigma(l(\zvec_s')))}
\mid^{-1}
\label{eq:vi_change}    
\end{equation}

\subsection{Evaluation} \label{sec:evaluation}
Evaluating the quality of the fitted variational distributions requires a comparison to the true posterior. In the case of low-dimensional problems, the two distributions can be compared visually.
In the case of higher-dimensional problems, this is not possible anymore. 
%
While the Evidence Lower Bound (ELBO) is a valuable metric for optimizing the parameters in VI, it is less helpful in comparing different approximations because it depends on the specific parametrization of the model \cite{yao2018yes}. Therefore,  \cite{yao2018yes} introduced $\hat{k}$ as a more suited approach for comparison, which since then has been used in other studies like \cite{dhaka2021challenges} to which we compare. 
The computation of $\hat{k}$ is based on the 
importance ratios which are defined as
\begin{equation}
    r_s = \frac{p(\thetavec_s, D)}{q_\lambdavec(\thetavec_s)} =  \frac{p(D \mid \thetavec_s) p(\thetavec_s) }{q_\lambdavec(\thetavec_s)}
\label{eq:weight} 
\end{equation}
If the variational distribution $q_\lambdavec(\thetavec)$ would be a perfect approximation of the posterior $p(\thetavec \mid D) \propto p(D \mid \thetavec) p(\thetavec)$, then important ratios $r_s$ would be constant. However, because of the asymmetry of the  KL-divergence used in the optimization objective (see Eq.~\ref{eq:vi1}), the fitted $q_\lambdavec(\thetavec)$ tends to have lighter tails than $p(\thetavec \mid D)$, with the effect that the distribution of $r_s$ are heavily right-tailed.  
To quantify the severity of the underestimated tails, a generalized Pareto distribution is fitted to the right tail of the $r_s$. The estimated shape parameter $\hat{k}$ of the Pareto distribution can be used as a diagnostic tool. A large $\hat{k}$ indicates a pronounced tail in the $r_s$ distribution and, hence, a bad posterior approximation. 
According to Yao et al., \cite{yao2018yes} values of $\hat{k} < 0.5$ indicates 
that the variational approximation $q_\lambdavec$ is good. Values of $0.5 < \hat{k} < 0.7$ indicate the variational approximation $q_\lambdavec$ is not perfect but still useful.  

\section{Experiments}
\label{sec:results}

We performed several experiments to benchmark our BF-VI approach versus exact Bayesian solutions, Gaussian-VI, and recent NF-VI approaches. All experiments were conducted using five repetitions, with the observed stability of Evidence Lower Bound (ELBO) optimization generally appearing independent of the randomly chosen starting values. Table~\ref{tab:models} shows an overview of the fitted models. The complete model definitions in Stan, along with the code for all experiments, can be found on GitHub.  


\begin{table*}[!ht]
\centering
\caption{%
Overview of the fitted Bayesian models in the benchmark experiments and the methods used to get the posteriors.  
MCMC or analytical solutions are used as ground truth, against which the quality of different VI approximations is compared (bold-faced methods outperform all others). The different VI approximations are either compared by the achieved $\hat{k}$ (see section \ref{sec:evaluation}) or visually by inspecting their deviation from the ground truth.
} \label{tab:models}
\resizebox{15cm}{!}{%
\begin{tabular}{@{}lll@{}}
\textbf{Experiment}             & \textbf{Model} & \textbf{Method (ground truth; VI  approximations - best is bold)}  \\ \hline{}
 Bernoulli       & ${Y \sim \Ber(\pi)}$   &  analytically; {\bf BF-VI},   MF-Gaussian-VI
\\
\hline{}
 Cauchy       & $Y \sim \text{Cauchy}(\xi, \gamma)$   &   MCMC; {\bf BF-VI},   MF-Gaussian-VI           \\
\hline{}
 Toy linear regression       & $(Y \mid x_1,x_2) \sim N(\mu_{\bf{x}}=\mu_0+\beta_1  x_1 + \beta_2  x_2, \sigma)$   &   MCMC; {\bf BF-VI},  MF-Gaussian-VI  \\
\hline{}
 Diamond       & $(Y \mid {\bf x}) \sim N(\mu_{\bf{x}}=\mu_0+\bf{x}^{\top}\bf{\beta},  \sigma)$  &   MCMC; BF-VI, {\bf MF-Gaussian-VI}, NVP-NF-VI, PL-NF-VI
\\
\hline{}
 NN non-linear regression       & $(Y \mid x) \sim N(\mu(x),\sigma)$  &    MCMC;  BF-VI,  MF-Gaussian-VI 
 \\
\hline{}
8schools      &   &   \\
 CP parametrization       & $(Y \mid\theta_n,\sigma_n) \sim N(\theta_n,\sigma_n)$, with $(\theta_n \mid \mu,\tau) \sim N(\mu,\tau)$,  
   &    MCMC; {\bf BF-VI}, MF-Gaussian-VI, NVP-NF-VI, PL-NF-VI          \\
   & \  $\mu \sim N(0,5)$, 
   $\tau \sim$ half-Cauchy$(0,5)$ & \\
 NCP parametrization      
 & $(Y \mid \mu, \tau, \tilde{\theta}_n,\sigma_n) \sim N(\mu+ \tau \cdot \tilde{\theta}_n, \sigma_n)$, with $\tilde{\theta}_n \sim N(0,1)$,  
   &     MCMC; {\bf BF-VI}, MF-Gaussian-VI, NVP-NF-VI, PL-NF-VI           \\
   &  \ 
   $\mu \sim N(0,5)$, 
   $\tau \sim$ half-Cauchy$(0,5)$ & \\
\hline{}
Melanoma      &   &   \\
 M1 (based on image $B$)      & $(Y \mid B) \sim \Ber\Big(\pi_{B}=\sigma\big(\mu(B)  \big)\Big)$  &   --; (Ensembling)           \\
 M2 (based on tabular feature $x$)      &  $(Y \mid x) \sim \Ber\Big(\pi_{x}=\sigma\big(\mu_0 + \beta_1 \cdot x  \big)\Big)$ &  MCMC; BF-VI 
     \\
 M3 (semi-structured using $x,B$ )     &  $(Y \mid B,x) \sim \Ber\Big(\pi_{(B,x)}=\sigma\big(\mu_0(B) + \beta_1 \cdot x  \big)\Big)$ &  --; BF-VI for $\beta_1$, non-Bayesian CNN for $\mu_0(B)$ 
\end{tabular}}
\end{table*}

\subsection{Models with a Single Parameter} 

First, we demonstrate with two single-parameter experiments that BF-VI can accurately approximate a skewed or bimodal posterior, which is impossible with Gaussian-VI. To obtain complex posterior shapes, we work with small data sets. 

\begin{figure}[h]
    \centering
    \includegraphics[width=0.45\textwidth]
    {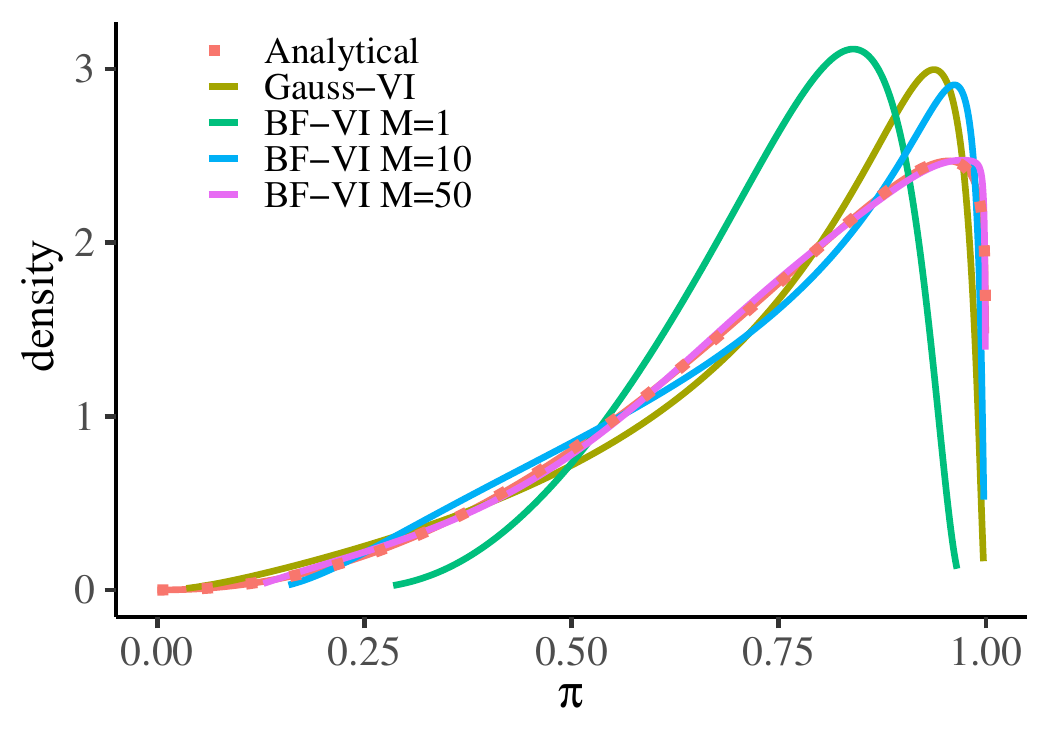}
     \includegraphics[width=0.45\textwidth]
    {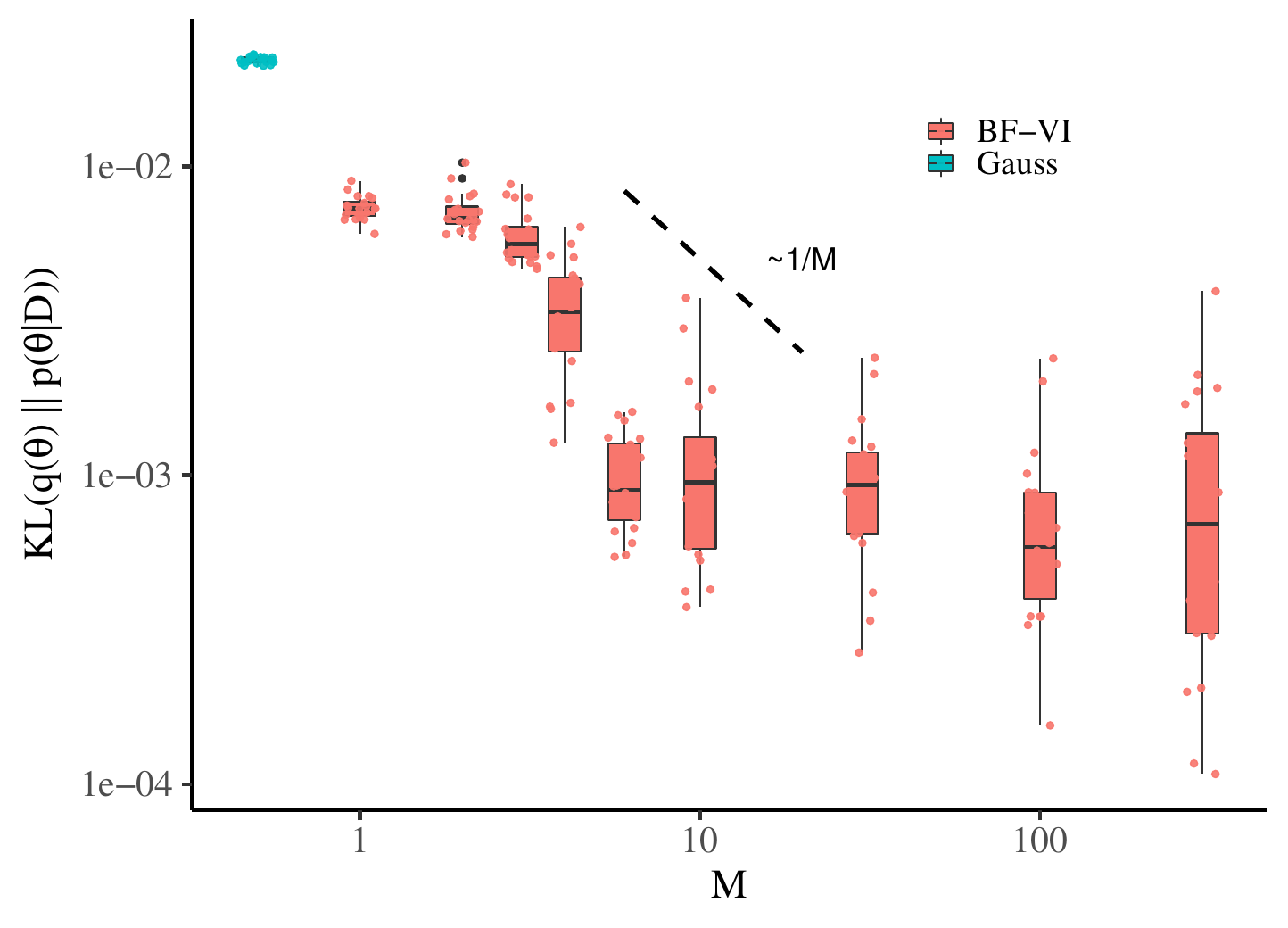}
    \caption{Bernoulli experiment. Left panel: Comparison of the analytical posterior for the parameter $\pi$ in the Bernoulli model ${Y \sim \Ber(\pi)}$ with variational distributions achieved via Gaussian-VI and BF-VI with BP order $M=1,10,50$. Right panel: The dependence of the  divergence $\dkl{q_\lambda(w)}{p(w \mid D)}$ on $M$ for 20 runs.}  
\label{fig:bernoulli}
\end{figure}

\begin{figure}[!h]
    \centering
    \includegraphics[width=0.90\columnwidth]
    {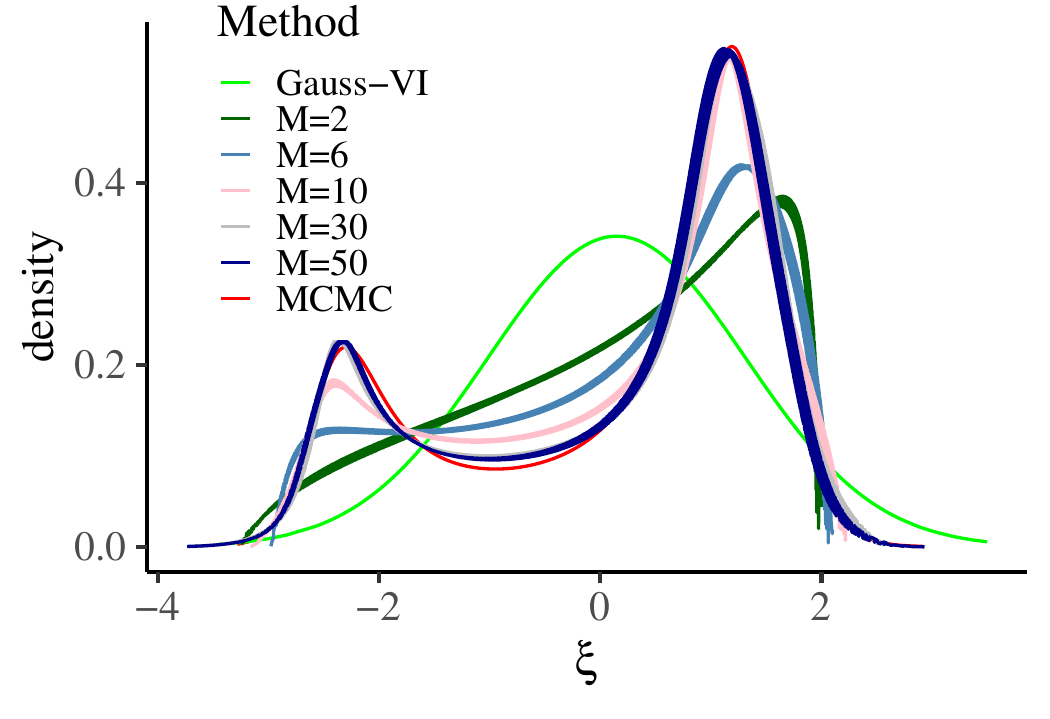}
    \caption{Cauchy experiment: comparison of MCMC posterior distribution of the parameter $\xi$ in the Cauchy model $Y\sim \text{Cauchy}(\xi, \gamma$) and the variational distributions estimated via Gaussian-VI or BF-VI ($M=2,6,10,30,50$). For the BF-VI method, the curves are overlays of 10 independent runs.} 
    \label{fig:cauchy}
\end{figure}

\paragraph{Bernoulli Experiment}
\ \\
We first look at an unconditional Bayesian model for a random variable $Y$ following a Bernoulli distribution ${Y \sim \Ber(\pi)}$ which we fit based on data $D$ consisting only of two samples ($y_1=1$, $y_2=1$). 
In this simple Bernoulli model, it is possible to determine the solution for the posterior analytically when using a beta distribution as prior. We choose $p(\pi)=\Be_{(1.1, 1.1)}(\pi) $ which leads to the conjugated posterior $p(\pi \mid D) = \Be_{(\alpha + \sum{y_i},\beta+n-\sum{y_i})}(\pi)$ (see analytical posterior in Fig.~\ref{fig:bernoulli}). 
We now use BF-VI to approximate the posterior. To ensure that the modeled variational distribution for $\pi$ is restricted to the support of $\pi\in[0,1]$, we pipe the result of the flow through an additional sigmoid transformation.
Fig.~\ref{fig:bernoulli} shows the achieved variational distributions after minimizing the negative ELBO and demonstrates the robustness of BF-VI: When increasing the order $M$ of the BP, the resulting variational distribution gets closer to the posterior up to a certain value of $M$ and then does not deteriorate when further increasing $M$. The left part of Fig.~\ref{fig:bernoulli} indicates a convergence order of $M$, which can also be proven for the one-dimensional case 
\ifappendix
(see Appendix \ref{sec:proof}).
\else
(proof not shown).
\fi
As expected, the Gaussian-VI does not have enough flexibility to approximate the analytical posterior nicely (see Fig.~\ref{fig:bernoulli}).
\paragraph{Cauchy Experiment}
\ \\
Here, we follow an example from \cite{yao2020stacking} and fit an unconditional Cauchy model $Y \sim \text{Cauchy}(\xi, \gamma)$ to six samples which we have drawn from a mixture of two Cauchy distributions $Y \sim \text{Cauchy}(\xi_1=-2.5, \gamma=0.5)+\text{Cauchy}(\xi_2=2.5, \gamma=0.5)$. Due to the misspecification of the model, the true posterior of the parameter $\xi$ has a bimodal shape which we have determined via MCMC (see Fig.~\ref{fig:cauchy}). 
We use BF-VI and Gaussian-VI to approximate the posterior of the Cauchy parameter $\xi$ by a variational distribution. 
As in the Bernoulli experiment, also here, BF-VI  has enough flexibility to accurately approximate the complex shape of the posterior when $M$ is chosen large enough. Further increasing $M$ does not deteriorate the approximation. Gauss-VI fails as expected.

\subsection{Models with Multiple Parameters} \label{sec:multi-parameter}
The following experiments use BF-VI in multi-parameter Bayesian models and benchmark the achieved solutions versus MCMC or published state-of-the-art VI approximations (see Table~\ref{tab:models}). In the following experiments, we did not tune the flexibility of our BF-VI approach but allowed it to be relatively high ($M=50$) since BF-VI does not suffer from being too flexible. Further, in all experiments in this section, we trained for $10^5$ epochs with 5 repetitions and set the number of samples for MC estimation to $S=10$ to be comparable with \cite{dhaka2021challenges}. From the repetitions and the posterior samples, we estimated $\hat{k}$ and the 90 $\%$ confidence intervals using the Rubins rule for the BF-VI and Gaussian-VI methods, with $S=50'000$ samples. Though runtime is not a consideration in this study, to provide context, a 2023 MacBook Pro's CPU processes approximately 100 epochs per second.   

\paragraph{Toy Linear Regression Experiment}
\ \\
To investigate if dependencies between model parameters are correctly captured, we use a simulated toy data set with two predictors and six data points to which we fit a Bayesian linear regression modeling the conditional outcome distribution $(Y \mid x_1,x_2) \sim \text{N}(\mu_{\bf{x}}=\mu_0+\beta_1  x_1 + \beta_2  x_2, \sigma)$.
\begin{figure}[!h]
    \centering
    \includegraphics[width=0.90\columnwidth]
    {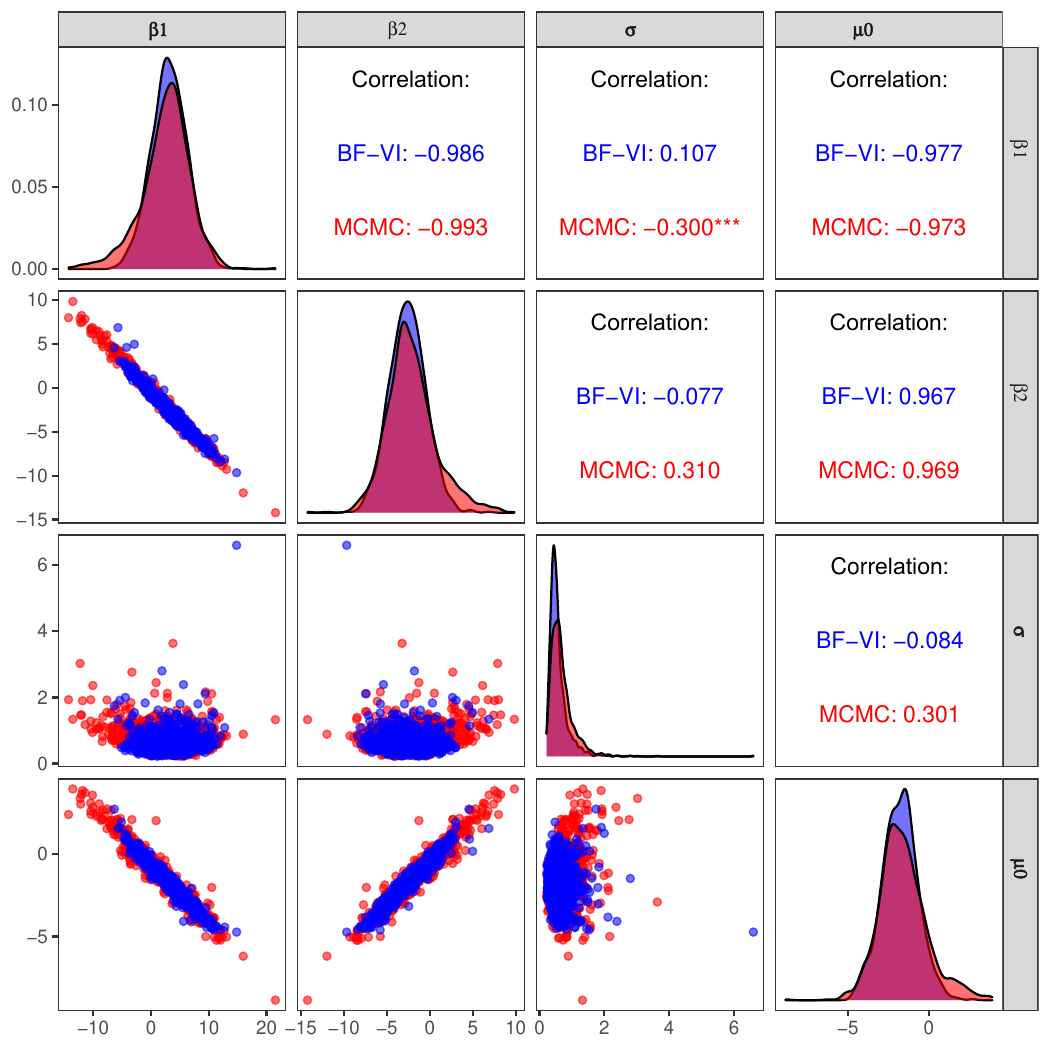}
    \caption{Toy linear regression example visualization of the posterior. The model has four parameters: two regression coefficients $\beta_1$ and $\beta_2$, the intercept $\mu_0$, and the standard derivation $\sigma$. Samples from the true posterior resulting from MCMC (red) are overlayed with samples from the BF-VI approximation (blue). }
    \label{fig:pairs}
\end{figure}

Fig.~\ref{fig:pairs} gives a visual impression of the joint true posterior of the four model parameters ($\mu_0, \beta_1, \beta_2, \sigma$) determined via MCMC samples (red) and its variational approximation (blue) achieved via BF-VI. The strong correlation between the regression coefficients ($\beta_1$, $\beta_2$) is nicely captured by the BF-VI approximation. Further, the skewness of posterior marginals involving \texttt{sigma}  is similar. However, we can see that BF-VI slightly underestimates the long tails of the posterior, confirming the known shortcoming of using the asymmetric KL-divergence in the objective function \cite{bleiVI}.  BF-VI ($\hat{k}=0.68 (0.51, 0.86)$) is superior to MF-Gaussian-VI ($\hat{k}=0.90 (0.77,1.01)$), which can not (by construction) capture the dependencies (MF) or the non-Gaussian shapes (see Fig.~\ref{fig:pairs.gauss}).

\paragraph{Diamond: Linear Regression  Experiment}
\ \\
The \textit{Diamond} linear regression benchmark example 
 $(Y \mid \xvec) \sim N(\mu_{x}=\mu_0 + \xvec^{\top}\betavec,  \sigma)$ has 26 model parameters and 5000 data points (see \cite{posteriordb} for reference MCMC samples and Stan code for a complete model definition).
Since we have much more data than parameters, the posterior is expected to be a narrow Gaussian around the maximum-likelihood solution, which is indeed seen in the MCMC solution \ifappendix
(see Fig.~\ref{fig:diamonds}).
\else
(data not shown).
\fi
In this setting, BF-VI or NF-VI cannot profit from their ability to fit complex distributions. Still, \cite{dhaka2021challenges} used this data set for benchmarking different VI methods, e.g., planar NF (PL-NF-VI), non-volume-preserving NF (NVP-NF-VI), and MF-Gaussian-VI. They achieved the best approximation via the simple Gaussian-VI  ($\hat{k}=1.2$).
The posterior approximation via PL-NF-VI and NVP-NF-VI have both been unsatisfactory ($\hat{k}=\infty$). We use the same amount of sampling ($S=10$) and achieve with BF-VI a better approximation of the posterior $\hat{k}=5.34(-2.52,13.20)$ but is still worse than the Gaussian-VI. A large spread in $\hat{k}$ indicates an unstable training procedure. This dataset is also quite challenging for MCMC simulations; we did not get satisfactory MCMC samples and took the reference posterior samples from the posteriorDB\footnote{https://github.com/stan-dev/posteriordb}.
\ifappendix
See Fig.~\ref{fig:diamonds} for a comparison of BF-VI and MCMC. 
\fi

\paragraph{8schools: Hierarchical Model  Experiment}
\ \\
The \textit{8schools} data set is a benchmark data set for fitting a Bayesian hierarchical model and is known to be challenging for VI approaches \cite{yao2018yes}, \cite{huggins2020validated}. It has 8 data points, corresponding to eight schools that have conducted independent coaching programs to enhance the SAT (Scholastic Assessment Test) scores of their students. 
There are two commonly used parameterizations of the model: centered parameterization (CP) and non-centered parameterization (NCP). NCP uses a transformed parameter to facilitate the MCMC sampling. See Table~\ref{tab:models} for more details of these parametrizations and \cite{posteriordb} for complete model definitions in Stan. 
In both parametrizations, the model has 10 parameters.
%
In \cite{dhaka2021challenges}, this benchmark data set was fitted with two NF-based methods and MF-Gaussian-VI. For the CP version \cite{dhaka2021challenges} 
$\hat{k}_{\text{CP}}=1.3, 1.1, 0.9$ was reported 
for PL-NF-VI, NVP-NF-VI, MF-Gaussian-VI respectively, which are all outperformed by our BF-VI method with $\hat{k}_{\text{CP}}=0.53 (0.11, 0.95)$. For the NCP version \cite{dhaka2021challenges} 
$\hat{k}_{\text{NCP}}=1.2, 0.7, 0.7$ was reported (same order), and again BF-VI  yields a superior $\hat{k}_{\text{NCP}}=0.36 (0.17, 0.55)$.
A visual inspection of the true MCMC posterior and its variational approximation again shows the underestimated distribution tails 
\ifappendix
(see Fig.~\ref{fig:8schools}). 
\else 
(data not shown).
\fi
For 8Schools and Diamond, a comparison with state-of-the-art models from the literature is summarized in Table~\ref{tab:khat}.
\begin{table}[h]
    \caption{Comparison posterior approximation with results from the literature for the Diamond and 8schools dataset in  CP and NCP parametrization. Shown is $\hat k$ (the lower, the better) for 
    different approximations: mean-Field Gaussian (MF-G) and Student-t (MF-T); NVP (NVP-VI) and Planar (P-VI) flow from \cite{dhaka2021challenges}, and BF-VI.}
    \label{tab:khat}
    \centering
    \begin{tabular}{l|c|c|c|c|c}
         & MF-G & MF-T & NVP-VI & P-VI& BF-VI  \\ \hline
         Diamond& {\bf 1.2} & 1.3 & $\infty$ & $\infty$ & 5.34(-2.52,13.20) \\ \hline
         8Schools (CP)& 0.9 & 0.9 & $1.3$ & $1.1$ & {\bf 0.53(0.11, 0.95)} \\ \hline
         8Schools (NCP)& 0.7 & 0.6 & $1.2$ & $0.7$ & {\bf 0.36(0.17, 0.55)} 
    \end{tabular}
\end{table}

\paragraph{NN-Based Non-linear Regression  Experiment}
\ \\
%
For this experiment, we use a small Bayesian NN for non-linear regression fitted on 9 data points. The model for the conditional outcome distribution is $(Y \mid \xvec) \sim N(\mu(\xvec),\sigma = 0.2)$. The small size of the used BNN, with only one hidden layer comprising 3 neurons and one neuron in the output layer giving  $\mu(\xvec)$, allows us to determine the posterior via MCMC. 
\begin{figure}[!h]
    \centering
    \includegraphics[width=0.90\columnwidth]
    {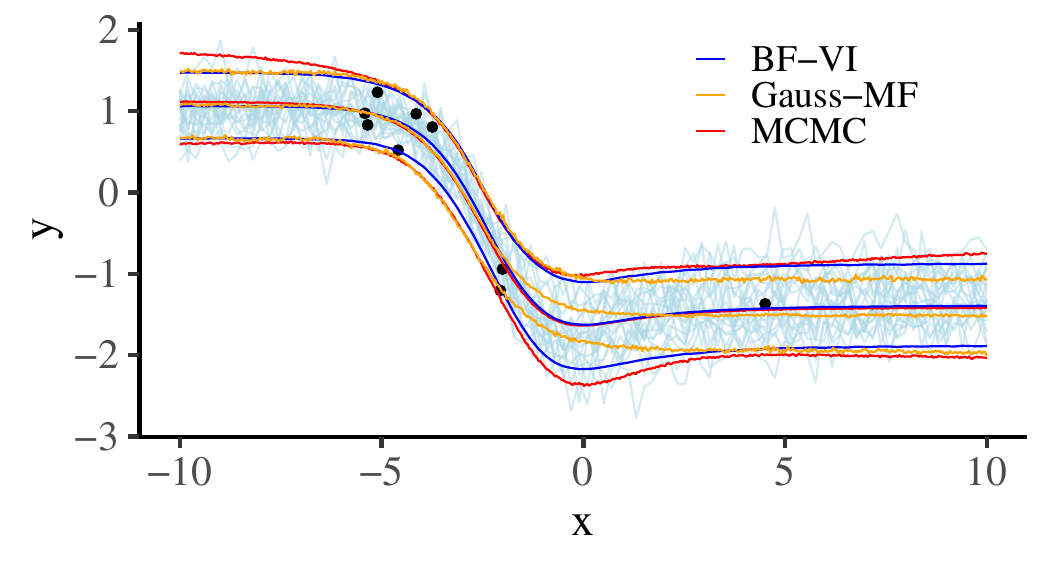} 
    \caption{Posterior predictive distribution of the non-linear regression model $(Y \mid\xvec) \sim N(\mu(\xvec),\sigma=0.2)$ where the conditional mean is modeled by a BNN using MCMC, MF-Gaussian-VI, or BF-FI. }
    \label{fig:dense}
\end{figure}
We then use  BF-VI and MF-Gaussian-VI to fit this BNN. Because the weights in a BNN with hidden layers are not directly interpretable, they are not of direct interest, and therefore, the fit of a BNN is commonly assessed on the level of the posterior predictive distribution  (see Fig.~\ref{fig:dense}). In this example, the more flexible BF-VI shows a slight improvement in approximating the true posterior predictive distribution when compared to the less complex approach with MF-Gaussian-VI, especially inside the regions where there is data (around $x=0$ in Fig.~\ref{fig:dense}). 

\subsection{Melanoma: Semi-structured NN Experiment}
\begin{figure}[!h]
    \centering
    \includegraphics[width=0.90\columnwidth]
    {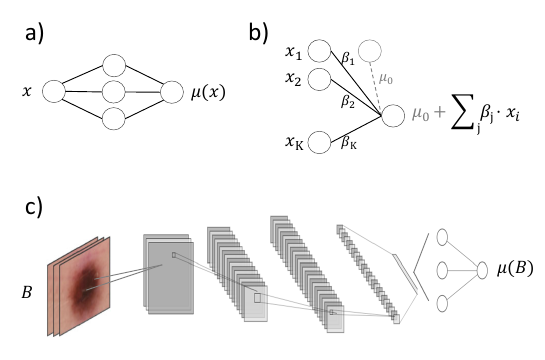} 
    \caption{The architecture of the used NN models or model parts. a) Dense NN with one hidden layer to model non-linear dependencies from the input (used in the NN-based non-linear regression example). b) Dense NN without hidden layer to model linear dependencies from tabular input data (used in M1 and M3 of the Melanoma experiment). c) CNN to model non-linear dependencies from the image input (used in M1 and M3).
    }
    \label{fig:nn}
\end{figure}

In this experiment, we use BF-VI  for semi-structured transformation models  \citep{kook_herzog2020} (see Fig.~\ref{fig:nn}), 
where complex data like images can be modeled by deep NNs and tabular data by interpretable model components. 
Please note that here, both the conditional distribution of the outcome 
$(y\mid B,x)$ and the unconditional posterior of the parameters are modeled by transformation models. Because of the deep NN model components involved, MCMC is not feasible anymore to determine the posterior. As a dataset, we use the SIIM-ISIC Melanoma Classification Challenge \footnote{\url{https://challenge2020.isic-archive.com}} data. The data comes from 33126 patients (6626 as test set, 21200 train, and 5300 validation set) with a confirmed diagnosis of their skin lesions, which is in $\approx 98$\% benign ($y=0$)  and in $\approx 2$\% malignant ($y=1$). The provided data $D=(B,x)$ is semi-structured since it comprises  (unstructured) image data $B$ from the patient's lesion along with (structured) tabular data $x$, i.e., the patient's age.

We fit the conditional outcome distribution 
$(Y \mid D) \sim \Ber\big(\pi_D)$ by modeling the probability for a lesion to be malignant $\pi_D=p(y=1 \mid D) = \sigma(h)$ applying  the sigmoid function $\sigma(\cdot)$ to a fitted transformation function $h: Y \rightarrow Z$. We study three models for $h$ depending on $x$ alone, $B$ alone, and in combination $B$ and $x$:\newline

{\bf M1 (DL-Model) $h=\mu(B)$:} As a baseline, we use a deep convolutional neural network (CNN) based on the melanoma image data 
(see Fig.~\ref{fig:nn} c) with a total of 419,489 weights to take advantage of the predictive power of DL on complex image data. For this DL model, we use deep ensembling \cite{lakshminarayanan2016simple} by fitting three CNN models with different random initializations and averaging the predicted probabilities.
The achieved test predictive performance and its comparison to other models are discussed in the last paragraph of this section. 

\begin{figure}[!h]
    \centering
    \includegraphics[width=0.90\columnwidth]
    {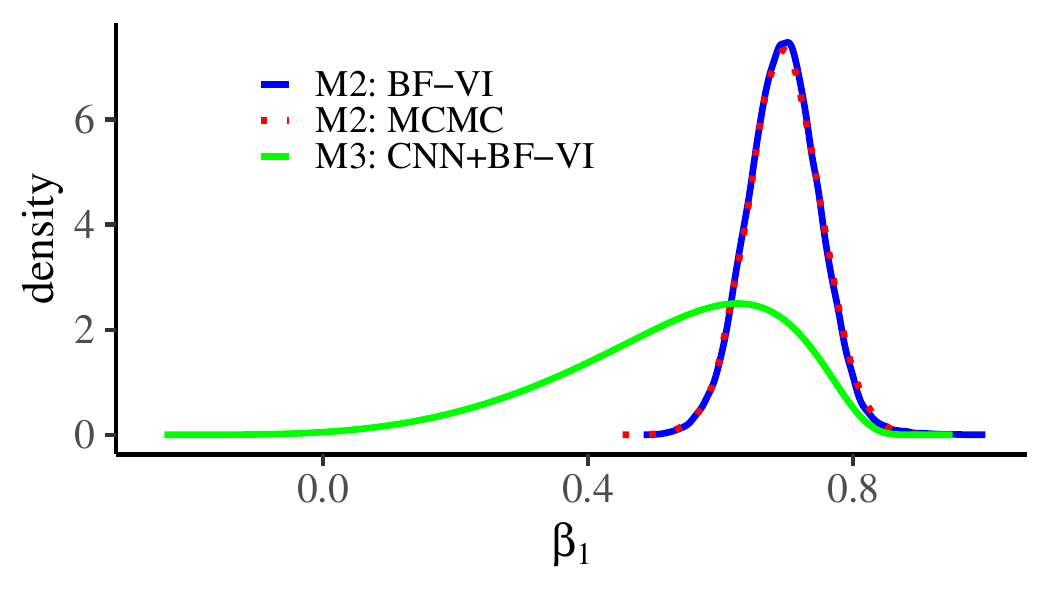} 
    \caption{Posteriors for the age-effect parameter $\beta_1$ in the Melanoma models M2 and M3.    }
    \label{fig:age}
\end{figure}

{\bf M2 (Logistic Regression) $h=\mu_0 +\beta_1 \cdot x$:}
When using only tabular features $x$, interpretable models can be built. We consider a Bayesian logistic regression with age as the only explanatory variable $x$ and use a BNN without a hidden layer to set up the model (see Fig.~\ref{fig:nn} b with only one input feature $x$). In logistic regression, a latent variable is modeled by a linear predictor $h=\mu_0 +\beta_1 \cdot x$, which determines the probability for a lesion to be malignant via $\pi_x = \sigma\big(\mu_0 +\beta_1 \cdot x\big)$ allowing to inter et $e^{\beta_1}$ as the odds-ratio, i.e., the factor by which the odds for lesions to be malignant changes when increasing the predictor $x$ by one unit. 
In Fig.~\ref{fig:age}, we compare the exact MCMC posterior of $\beta_1$  with the BF-VI approximation, demonstrating that BF-VI accurately approximates the posterior.

{\bf M3 (semi-structured) $h=\mu(B) +\beta_1 \cdot x$:} 
This model integrates image and tabular data and combines the predictive power of M1 with the interpretability of M2. 
We use a (non-Bayesian) CNN that determines $\mu(B)$ and BF-VI for the NN without a hidden layer that determines $\beta_1$ (see Fig.~\ref{fig:nn} b and c).  Both NNs are jointly trained by optimizing the ELBO. The resulting posterior for $\beta_1$ differs from the simple logistic regression (see Fig.~\ref{fig:age}), indicating a diminished effect of age after including the image. 
 Again, $e^{\beta_1}$ can be interpreted as the factor by which the odds for a  lesion to be malignant change when increasing the predictor age by one unit and holding the image constant.  

While the main interest of our study is on the posteriors, we also determine the predictive performance on the test set. 
To quantify and compare the test prediction performances, we look at the achieved log scores
(M1: -0.076, M2: -0.085, M3: -0.076) and the AUCs with 95$\%$ CI (M1: $0.83 (0.79, 0.86)$, M2: $0.66 (0.61, 0.71)$, M3: $0.82 (0.79, 0.85)$). 
For both measures, higher is better. Interestingly, the image-based models (M1, M3) have higher predictive power than M2, which only uses tabular data. The semi-structured model M3, including tabular and image information, has a similar predictive power compared to M1, which only uses images. The benefit of the semi-structured model here is that it provides interpretable parameters for the tabular data along with uncertainty quantification without losing predictive performance. 
\section{Summary and Outlook} 
\label{sec:outlook}
The proposed BF-VI is flexible enough to approximate any posterior in principle without being restricted to variational distributions with known parametric distribution families like Gaussians. In benchmark experiments, BF-VI accurately fits non-trivial posteriors in low-dimensional problems in a BBVI setting. 
For higher dimensional models, BF-VI outperforms published results from other NF-VI methods \cite{dhaka2021challenges} on the studied benchmark data sets. 
Still, we observe that the posterior can not be fitted perfectly in high dimensions by BF-VI, especially since the tails of the approximation are too short. We attribute this limitation to known difficulties in the optimization process and the asymmetry of the KL divergence. These challenges of VI were not in the focus of our study, and we leave it to further research.

To the best of our knowledge, we are the first to demonstrate how BBVI can be used in semi-structured models. We used BF-VI on the public melanoma challenge data set, integrating image data and tabular data by combining a deep CNN and an interpretable model part. 
We see a valuable application of BF-VI in models with interpretable parameters, i.e., statistical or semi-structured models where we can model complex posterior distributions of the interpretable parameters. 
Especially in semi-structured models with deep NN components that cannot be fitted with MCMC, BF-VI allows determining the variational distribution for the interpretable model parts. Moreover, efficient SGD optimizers can be used in BF-VI to fit all model parts jointly. We plan to extend our research on BF-VI for semi-structured models in the future and investigate the quality of the posterior approximations.


\section{Acknowledgements}
\label{sec:acknowledgements}
 The research of BS was supported by the Novartis Research Foundation (FreeNovation~2019). The research of the OD and DD was partly supported by the Federal Ministry of Education and Research of Germany	(BMBF) in the project DeepDoubt (grant no. 01IS19083A). We further, would like to thank Nadja Klein, Lucas Kook, and Rebekka Axthelm for fruitful discussions. 

\section*{Conflict of Interest}
None

\bibliographystyle{IEEEtran}
\bibliography{IEEEabrv, main}

\ifappendix
\appendix
\newpage
\renewcommand{\thepage}{S\arabic{page}} 
\renewcommand{\thesection}{S\arabic{section}} 
\renewcommand{\thesection}{\Alph{section}} 
\renewcommand{\thetable}{S\arabic{table}}  
\renewcommand{\thefigure}{S\arabic{figure}}
\renewcommand{\theequation}{S\arabic{equation}}
\setcounter{table}{0}
\setcounter{figure}{0}
\setcounter{equation}{0}
\setcounter{section}{0}
\onecolumn
\section*{Appendix}
\ifproof
\section{Tightness of Variational Bound in the Limit $M \rightarrow \infty$, \label{sec:proof}}
In the following, we show for the one-dimensional case that 
for a transformation function built from Bernstein polynomials of order $M$
there exists a solution
so that \dkl{q_{\lambda,M}(\theta)}{p(\theta  \mid D)} in  (\ref{eq:vi1}) asymptotically vanishes with $1/M$. 

We assume that the posterior $p(\theta \mid D)$ is continuous in the compact interval $[\theta_a, \theta_b]$ with CDF $F_\theta(\theta)$ and that the first three derivatives of $F_\theta(\theta)$ exists. Without loss of generality, we set $\theta_a=0$ and $\theta_b=1$. This allows us to construct a transformation $h:\theta \rightarrow Z$ transforming between  $\theta$ with CDF $F_\theta$ and  a random variable $Z$ with CDF $F_Z$ by requiring $F_Z(h(\theta)) = F_{\theta}(\theta)$ via 
\begin{equation}
    h(\theta) = F_Z^{-1}(F_{\theta}(\theta)).
    \label{eq:h_theta}
\end{equation}
 We assume that $F_Z$ has continuous derivatives up through order 3, then from  (\ref{eq:h_theta}), it follows that $h(\theta)$ has continuous derivatives up to order 3. Further, we require that the number of roots of $h''(\theta)$ and $h'''(\theta)$ are countable. In addition, we require that the density $p_Z(z)>0$. We approximate the posterior density $p(\theta \mid D)$ by $q_\lambda(\theta)$ by approximating $h(\theta)$ with a Bernstein polynomial of order $M$ given by:
\begin{equation}
    B_M(\theta) 
    = \sum_{m=0}^{M} h\left(\frac{m}{M}\right) 
    \binom{M}{m}
    \; \theta^m (1-\theta)^{M-m}
    \label{eq:bern_poly}
\end{equation}
Note that we fixed the coefficients of the transformation here. This is valid since we just want to prove that a particular solution exists for that (\ref{eq:vi1}) asymptotically vanishes with $1/M$. As shown in \cite{farouki2012bernstein}, the Bernstein approximation "is at least as smooth" as $h(\theta)$, in our case that at least the first 3 derivatives of  $B_M(\theta)$ exists and converge to the corresponding derivatives of $h(\theta)$. In a first step, we upper bound the KL-divergence to that part of $\theta' \subset \theta $ where $\log(\frac{q_\lambda(\theta)}{p(\theta \mid D)})>0$ since  $q_\lambda(\theta) \ge 0$, the following approximation hold
\begin{equation*}
\dkl{q_\lambda(\theta)}{p(\theta \mid D)} = 
\int_0^1 q_\lambda(\theta) 
\log
\left(
\frac{q_\lambda(\theta)}{p(\theta \mid D)}
\right)
\;d\theta
\le
\int_{\theta'}q_\lambda(\theta) 
\log
\left(
\frac{q_\lambda(\theta)}{p(\theta \mid D)}
\right)
\;d\theta
\end{equation*}
Using the change of variable formula, we can express the densities of the posterior as 

$p(\theta \mid D)=p_Z\left(h(\theta)\right) \cdot  \lvert h'(\theta) \rvert $ and its approximation as $q_\lambda(\theta) = p_Z\left(B_M(\theta)\right)  \cdot \lvert B'_M(\theta)\rvert$, the dash indicates a derivative w.r.t. $\theta$. Hence, the KL-Divergence is bounded by:
\begin{equation}
\dkl{q_\lambda(\theta)}{p(\theta \mid D)} \le
\int_{\theta'} q_\lambda(\theta) 
\log\left(
\frac{p_Z\left(B_M(\theta)\right)  \cdot \lvert B'_M(\theta)\rvert}
{p_Z\left( h( \theta )\right) \cdot  \mid h'(\theta ) \mid} 
\right)
\;d\theta
\label{eq:kl_b1}
\end{equation}
The transformation function $h(\theta)$ is either strictly monotonic increasing or decreasing, w.l.o.g. we assume that $h'(\theta) > 0$. This leads to  ordered coefficients $h(m/M)$  and so $B'_M(\theta) > 0$.

Since the density $p_Z(z)$ and the derivative $B'_M(\theta)$ of Bernstein approximation are finite, so is $q_\lambda(\theta) = p_Z\left(B_M(\theta)\right)  \cdot  \mid B'_M(\theta) \mid$. This allows to upper bound the integral in (\ref{eq:kl_b1}) with $C=\sup(q_\lambda(\theta))$ in the range $\theta \in [0,1]$  by:
\begin{equation*}
    \dkl{q_\lambda(\theta)}{p(\theta \mid D)} \le
    C \cdot \underbrace{\int_{\theta'} 
    \log\left(
    \frac{p_Z\left(B_M(\theta)\right)}
    {p_Z\left(h(\theta)\right) } 
    \right) \;d\theta}_{=:I_1}
    + C \cdot
     \underbrace{
     \int_{\theta'} 
    \log\left(
    \frac{ B'_M(\theta)}
    { h'(\theta)} 
    \right) \;d\theta}_{=:I_2}
\end{equation*}
With increasing $M$ the term $\Delta = B_M(\theta) - h(\theta)$ gets arbitrarily small. We do a Taylor expansion $p_Z(B_M(\theta)) = p_Z(h(\theta) + \Delta) = p_Z(h(\theta)) + p'_Z(h(\theta)) \Delta + \mathcal{O}(\Delta^2)$. With $\log(1+x) \le x$ the first integral $I_1$ can be approximated as
\begin{equation*}
    I_1 = \int_{\theta'}
    \log\left(
    1 + \frac{p_Z'\left(h(\theta)\right)}{p_Z\left(h(\theta)\right)} \Delta + \mathcal{O}(\Delta^2) 
    \right) \;d\theta
    \le
    \int_{\theta'}
    \left(
    \frac{p_Z'\left(h(\theta)\right)}{p_Z\left(h(\theta)\right)} \Delta + \mathcal{O}(\Delta^2) 
    \right) 
    \;d\theta
\end{equation*}
Since $p_Z(z) > 0$ we can set $C_2=\sup(\frac{p_Z'\left(h(\theta)\right)}{p_Z\left(h(\theta)\right)})$ yielding:
\begin{equation*}
    I_1 
    \le
    C_2 \cdot \int_{\theta'}  \left(B_M(\theta) - h(\theta) \right) \;d\theta + \mathcal{O}(\Delta^2)  
\end{equation*}
The Voronovskaya theorem (see \cite{farouki2012bernstein}) states that $B_M(\theta)-h = \theta/(2 M) h''(\theta) + o(M^{-1})$ for $ h''(\theta) \ne 0$. Assuming that $ h''(\theta) = 0$ only on a countable set of points, the asymptotic of $I_1$ is given by $1/M$. 

Using $\log(x) \le x - 1$, the second integral can be bounded as follows. 
\begin{equation*}
    I_2 = 
    \int_{\theta'}
    \log\left(
    \frac{ B'_M(\theta)}
    { h'(\theta)} 
    \right) \;d \theta
    \le
   \int_{\theta'}
    \left(
    \frac{ B'_M(\theta)}
    { h'(\theta)} 
    - 1
    \right) \;d \theta
    = 
   \int_{\theta'}
    \frac{1}{h'(\theta)}
    \left(
     B'_M(\theta)
     -  h'(\theta)
    \right) \;d \theta
\end{equation*}
According to our assumptions $h'(\theta) > 0$ and so the supremum $C_3=\sup(1/h'(\theta))$ exists. 
\begin{equation*}
    I_2 
    \le
    C_3
   \int_{\theta'}
    \left(
     B'_M(\theta)
     -  h'(\theta)
    \right) \;d \theta
    \le 
    \frac{1}{M}
    \int_{\theta'}
    \left(
    \theta/2 h'''(\theta) + 
    1/2  h''(\theta) + 
    o(M^{-1})
    \right) \; d\theta
\end{equation*}
 This shows that the KL-Divergence in  (\ref{eq:vi1}) asymptotically converges to zero as $1/M$. 
\fi

\section{Experimental Details of and Additional Results}
In the following, we give additional details for the experiments. For the examples using MCMC simulations as a benchmark, the Stan-code can be found at  \url{https://github.com/tensorchiefs/bfvi_paper/tree/main/mcmc}. If not otherwise stated, the training has been as described in the main text. If not given below, the used data can be found at: \url{https://github.com/tensorchiefs/bfvi_paper/tree/main/data}.   
\subsection{Bernoulli Example}
\subsubsection*{The Data Generation Process}
Two samples are drawn from a Bernoulli distribution.

\subsubsection*{The Actual Data}
The realizations are $y_1 = 1$ and $y_2 = 2$.

\subsubsection*{Details of the Training Procedure}
To demonstrate the possibility that the described solution converges in principle to the exact solution, we used a large number of MC samples $S=2500$ and epochs=2500 for the training. 
Figure \ref{fig:loss_bernoulli} shows the loss curve for different values of $M$ ($M=10$, $30$, $50$, and $100$), along with Gauss-VI for comparison.

\begin{figure}[h]
  \centering
  \includegraphics[width=0.8\linewidth]{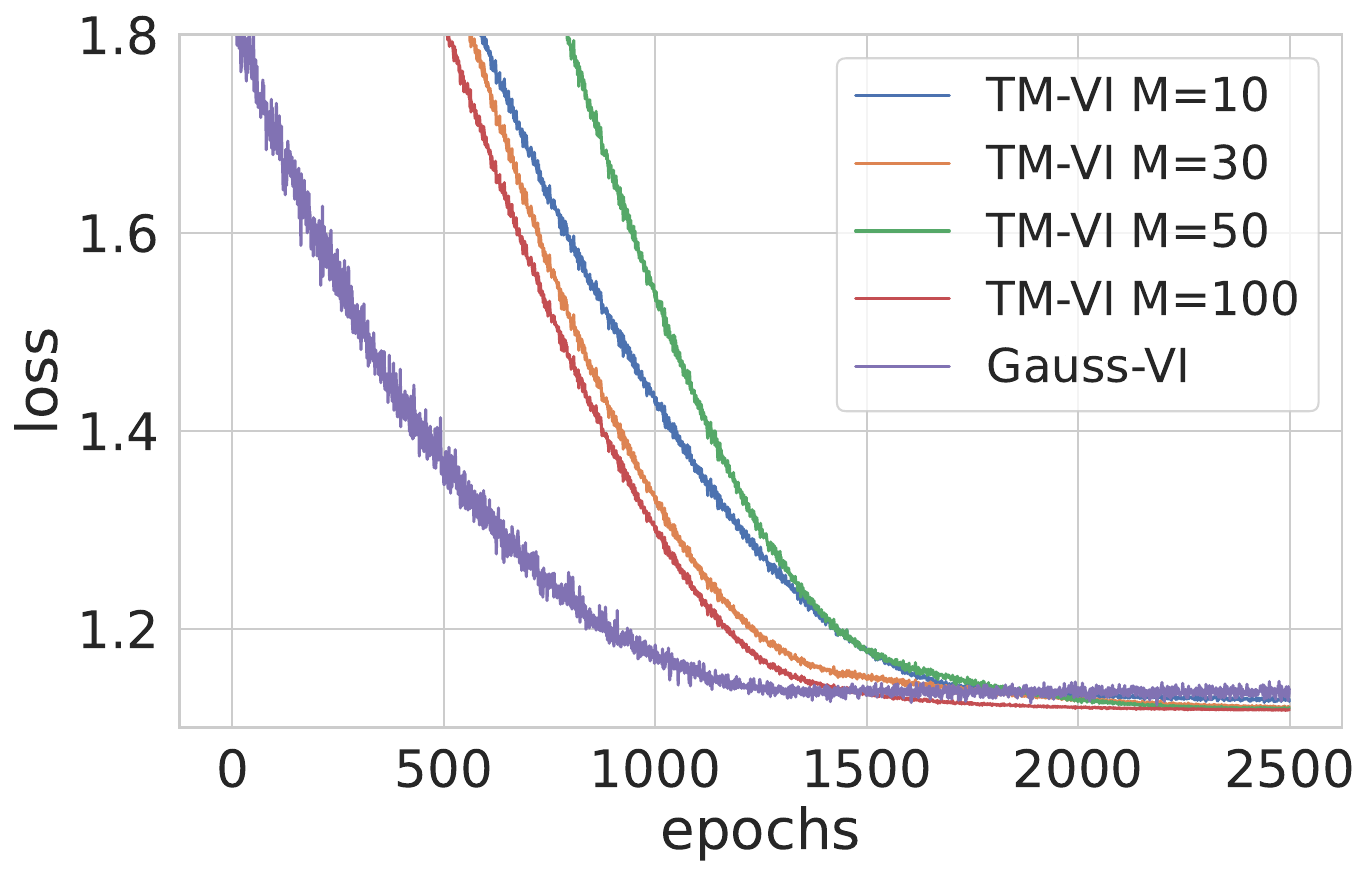}
  \caption{The loss ELBO in  the Bernoulli Example for $M=10$, $30$, $50$, $100$, and Gauss-VI.}
  \label{fig:loss_bernoulli}
\end{figure}
\subsubsection{Additional Results (Tightness of the ELBO)}
For this example, we can estimate the KL divergence via

\begin{equation*}
\dkl{q_\lambda(\theta)}{p(\theta \mid D)} =
\int~q_\lambda(\theta)\log\left(\frac{q_\lambda(\theta)}
{p(\theta \mid D)}\right)d\theta \\  \approx
\frac{1}{S} \sum_{\theta_s \sim q_\lambda}  \log\left(\frac{q_\lambda(\theta_s)} {p(\theta_s \mid D)}\right) 
\label{eq:kl_1}
\end{equation*}
The sum is calculated from the samples $\theta_s$ of the trained approximative distribution, which can be obtained via the trained transformation function. The likelihood $p(D \mid\theta_s)$ and prior $p(\theta_s)$ for those samples can be easily calculated and probabilities $q_\lambda(\theta_s)$ can be calculated via (\ref{eq:vi_change}). In this simple case, the posterior can be calculated analytically to $p(\theta_s  \mid D)  = \Be_{(\alpha + \sum{y_i},\beta+n-\sum{y_i})}(\theta_s) =  \Be_{(3.1, 1.1)}(\theta_s)$.  

In Fig.~\ref{fig:bernoulli}, we show a direct comparison of the posterior and its approximation of the dependence of the tightness of the ELBO on $M$, importantly increasing $M$ and thus the flexibility does not deteriorate the approximation.

\subsection{Cauchy Example}\label{par:cauchy}
\subsubsection*{The Data Generation Process}
Six samples are drawn from a mixture of two Cauchy distributions $y \sim \text{Cauchy}(\xi_1=-2.5, \gamma=0.5)+\text{Cauchy}(\xi_2=2.5, \gamma=0.5)$.

\subsubsection*{The Actual Data}
From the data generating process, we draw the following 6 examples $y = 1.2083935,-2.7329216,  4.1769943,  1.9710574, -4.2004027, -2.384988$. 

\subsubsection*{Details of the Model}
We use an unconditional Cauchy model $y \sim \text{Cauchy}(\xi, \gamma = 0.5)$ were $\gamma=0.5$ is given. Note that this model is misspecified and gives rise to a multimodal posterior. The location parameter $\xi$ is assigned a standard normal prior $\xi \sim N(0,1)$.
\subsubsection*{Details of the Training Procedure}
To demonstrate the possibility that the described solution converges in principle to the exact solution, we used a rather large number of MC samples $S=1000$ and epochs=1000 for the training.  
Figure \ref{fig:loss_Cauchy} shows the loss curve (-ELBO) for different values of $M$ ($M=1$, $M=10$, $30$, and $50$), along with Gauss-VI for comparison. 

\begin{figure}[h]
  \centering
  \includegraphics[width=0.8\linewidth]{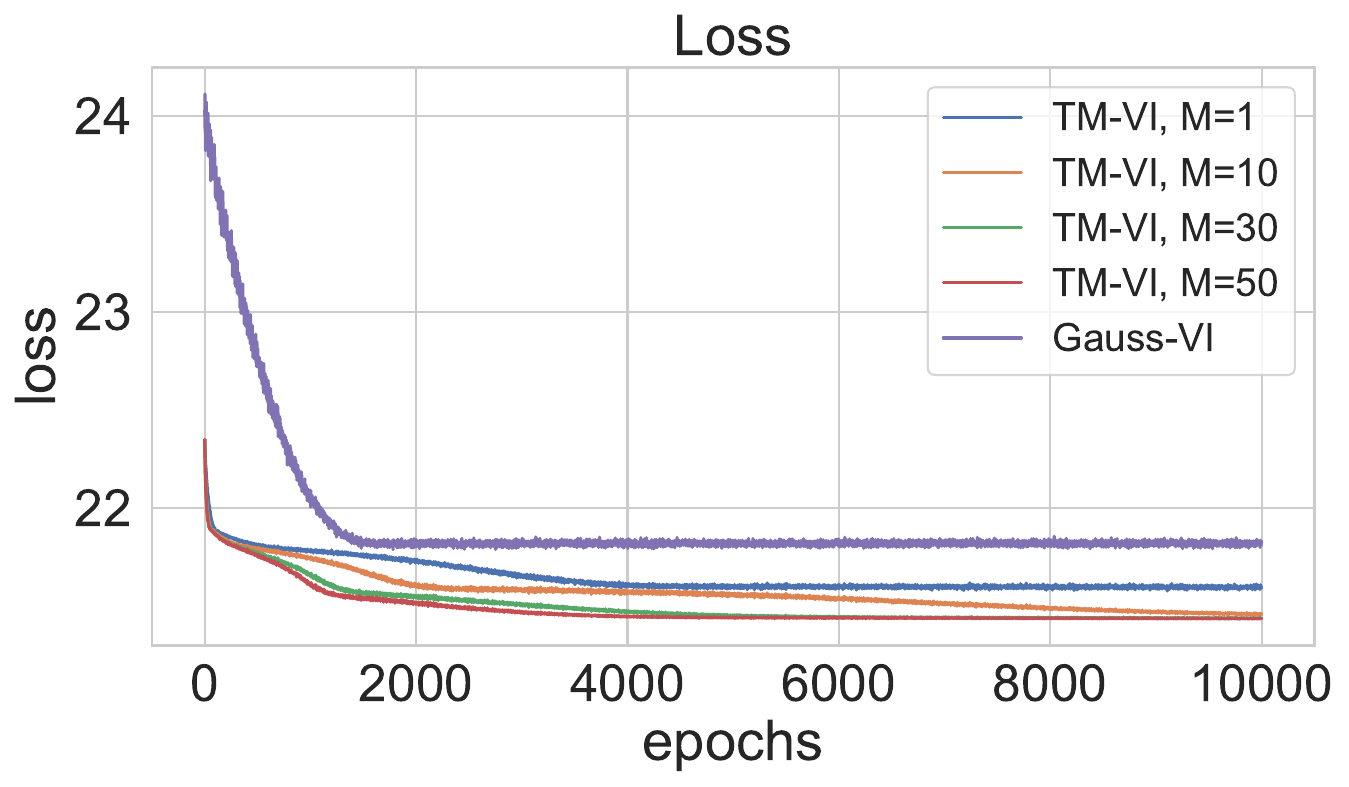}
  \caption{The loss curve for $M=1$, $M=10$, $30$, $50$ and Gauss-VI for the Cauchy example.}
  \label{fig:loss_Cauchy}
\end{figure}

\subsubsection{Additional Results: Tightness of the ELBO}
For this example, we cannot determine the posterior $p(\theta_s \mid D) = \frac{p(D  \mid \theta_s) p(\theta_s)}{\int p(D  \mid \theta_s) p(\theta_s) \; d\theta}$ analytically anymore. However, we can rewrite (\ref{eq:kl_1}) to:
\begin{equation}
\begin{split}
\dkl{q_\lambda(\theta)}{p(\theta \mid D)} 
&\approx \frac{1}{S} \sum_{\theta_s \sim q_\lambda}  \log\left(\frac{q_\lambda(\theta_s)} {p(\theta_s \mid D)}\right) \\
&= \frac{1}{S} \sum_{\theta_s \sim q_\lambda}  \log\left(\frac{q_\lambda(\theta_s)} {p(D \mid \theta_s) p(\theta_s)}\right) 
+ \underbrace{\log\left( \int p(D  \mid \theta) p(\theta) d\theta \right)}_{=\log(P(D))}
\end{split}
\label{eq:kl_2}
\end{equation}
Numerical integration obtains the log-evidence $\log(P(D)) =-21.43069$ for this example.

In Fig.~\ref{fig:cauchy_abla}, we show a direct comparison of posteriors. The dependence of the tightness of the ELBO on $M$ calculated in the lower part is calculated via (\ref{eq:kl_2}).

\subsubsection{Additional Results: Effect of Pre-Processing}\label{appendix:trafo_design}
In Fig.~\ref{fig:cauchy_abla}, the effect of different transformations before the Bernstein polynomials is studied.

\begin{figure}[!h]
    \centering
    \includegraphics[width=0.90\textwidth]
    {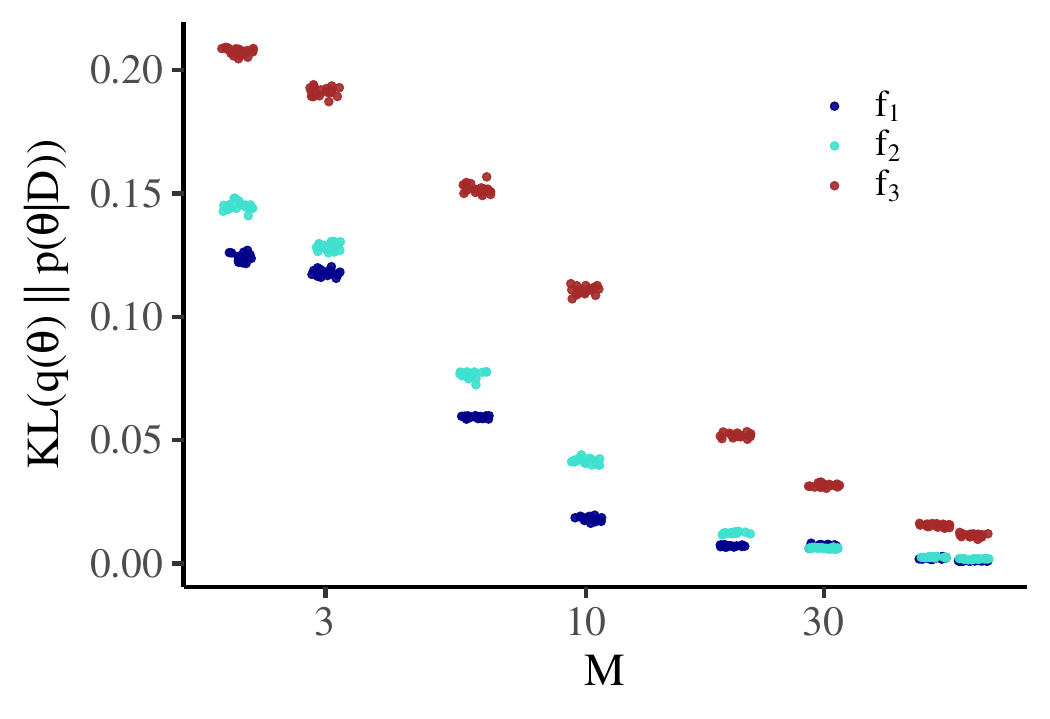}
    \caption{Effect of the different transformations before the $f_{\text{BP}}$ on the performance. In $f_1$, $f_2$ $z \sim N(0,1)$ and the transformation are $f_1 = f_\text{BP} \circ \sigma \circ l$ and $f_2 = f_\text{BP} \circ \sigma$. For $f_3$, $z$ is sampled from a truncated normal with shape 0.5, scale 0.15, and with bound $10^{-4}$ to $1-10^{-4}$ and transformed via $f_\text{BP}$.}    
    \label{fig:cauchy_abla}
\end{figure}

There are different options to ensure that the input to BP fulfills the restriction $z \in [0,1]$. One option is to use a latent distribution with unrestricted. A random variable $z'$ from that distribution is then "piped" through a sigmoid function $\sigma:[-\infty, \infty] \rightarrow [0,1]$. Note that the total transformation $f_\text{BP} \circ \sigma$ stays monotonically increasing. This construction allows us to choose 
the Standard-Normal for $F_{Z'}$. We have observed in low-dimensional examples a more efficient training and better fitting with fewer parameters $M$  when additionally prepending an affine transformation function $l = \alpha \cdot z + \beta$ before $\sigma$ so that the total transformation is $f=f_\text{BP} \circ \sigma \circ l$ (see section \ref{par:cauchy}). 
We also experimented with an additional affine function after the Bernstein polynomials but did not find a beneficial effect. We use $f=f_\text{BP} \circ \sigma \circ l$ in the following, except in \ref{par:cauchy} where we studied other possibilities to restrict the input to $f_\text{BP}$ to $[0,1]$. Moreover, a distribution $F_Z$ with support $[0,1]$ can be used. We experimented with a truncated Gaussian. 

 While a for large number $M$ of the performance for the different methods become similar (see Fig.~\ref{fig:cauchy_abla}), the transformation $f_1 = f_\text{BP} \circ \sigma \circ l$ works better for smaller of $M$, we thus choose this transformation for all experiments.

\subsection{Toy Linear Regression Example}\label{par:toy}

In this example, a posterior is generated, which displays a strong correlation between the results.
\subsubsection*{The Data Generation Process}
The data is generated by the following R-code.
\begin{lstlisting}[language=R]
  set.seed(42)
  N = 6L # number data points
  P = 2L #number of predictors
  z_dg = rnorm(N, mean=0, sd=1)
  x_dg = scale(rnorm(N, mean = 1.42*z_dg, sd = 0.1), 
               scale = FALSE)
  y_r = rnorm(N, mean = 0.42*x_dg - 1.1 * z_dg, sd=0.42)
\end{lstlisting}
\subsubsection*{The Actual Data}
\begin{lstlisting}[language=R]
>   x_r
           [,1]        [,2]
[1,]  1.3709584  1.48475156
[2,] -0.5646982 -1.42449894
[3,]  0.3631284  0.10432308
[4,]  0.6328626  0.27923186
[5,]  0.4042683  0.09138635
[6,] -0.1061245 -0.53519391
>   y_r
[1] -1.46778013 -0.09421285 -0.41162052 
-0.31177232 -0.52569912 -1.22375575
\end{lstlisting}

\subsubsection*{Details of the Model}
The model is a linear regression with Gaussian noise, defined by the likelihood \( y \sim \mathcal{N}(x w + b, \sigma) \). The priors are specified as \( b \sim \mathcal{N}(0, 10) \), \( w \sim \mathcal{N}(0, 10) \), and \( \sigma \sim \text{Lognormal}(0.5, 1) \).

\subsubsection*{Details of the Training Procedure}
Figure \ref{fig:loss_all} presents the scaled loss curves for the first run of the remaining experiments. Plotted are the epochs versus the negative ELBO, where the curve is normalized by first subtracting the minimum value observed across all epochs and then dividing by the absolute value of this minimum. For the remaining 4 runs, similar behavior is observed. To be comparable with existing literature the ELBO is estimated with $S=10$ MC samples introducing some level of noise into the data. 

\subsubsection{Additional Results}
Figure \ref{fig:pairs.gauss}, shows the result for MF-Gaussian approximation. As is visible, the mean-field approach cannot capture the multivariate correlation of the posterior and strongly focuses on the mode of the posterior.
\begin{figure}[!h]
\begin{center}
    \includegraphics[width=0.90\textwidth]
    {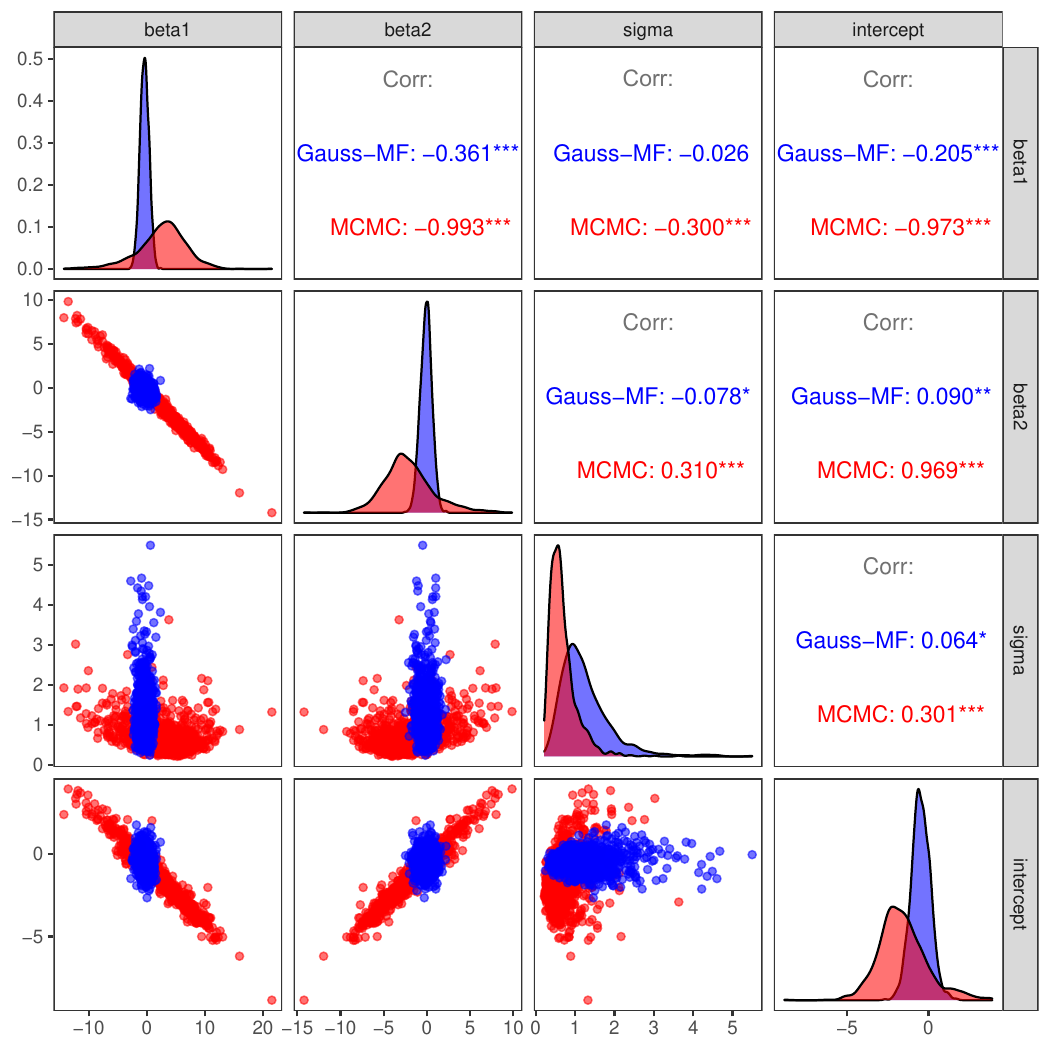}
    \caption{Mean-Field Gaussian results for the linear regression posteriors.}
    \label{fig:pairs.gauss}
    
\end{center}
\end{figure}

\subsection{Diamond}
The diamond data set is modeled with a linear model. To be in accordance with \cite{posteriordb}, we choose standard normal priors, except for $\sigma$ and the intercept, for which we choose Student's t-distributions as priors. Specifically, the intercept is assigned a Student's t-distribution with 3 degrees of freedom, a location parameter of 8, and a scale parameter of 10, while $\sigma$ is modeled using a half-Student's t-distribution with 3 degrees of freedom, a location of 0, and a scale parameter of 10.

\subsubsection{Additional Results}
In Fig.~\ref{fig:diamonds}, we show samples from the marginals of the MCMC solution and the BF-VI approximation. 
\begin{figure}[!h]
    \includegraphics[width=0.90\textwidth] {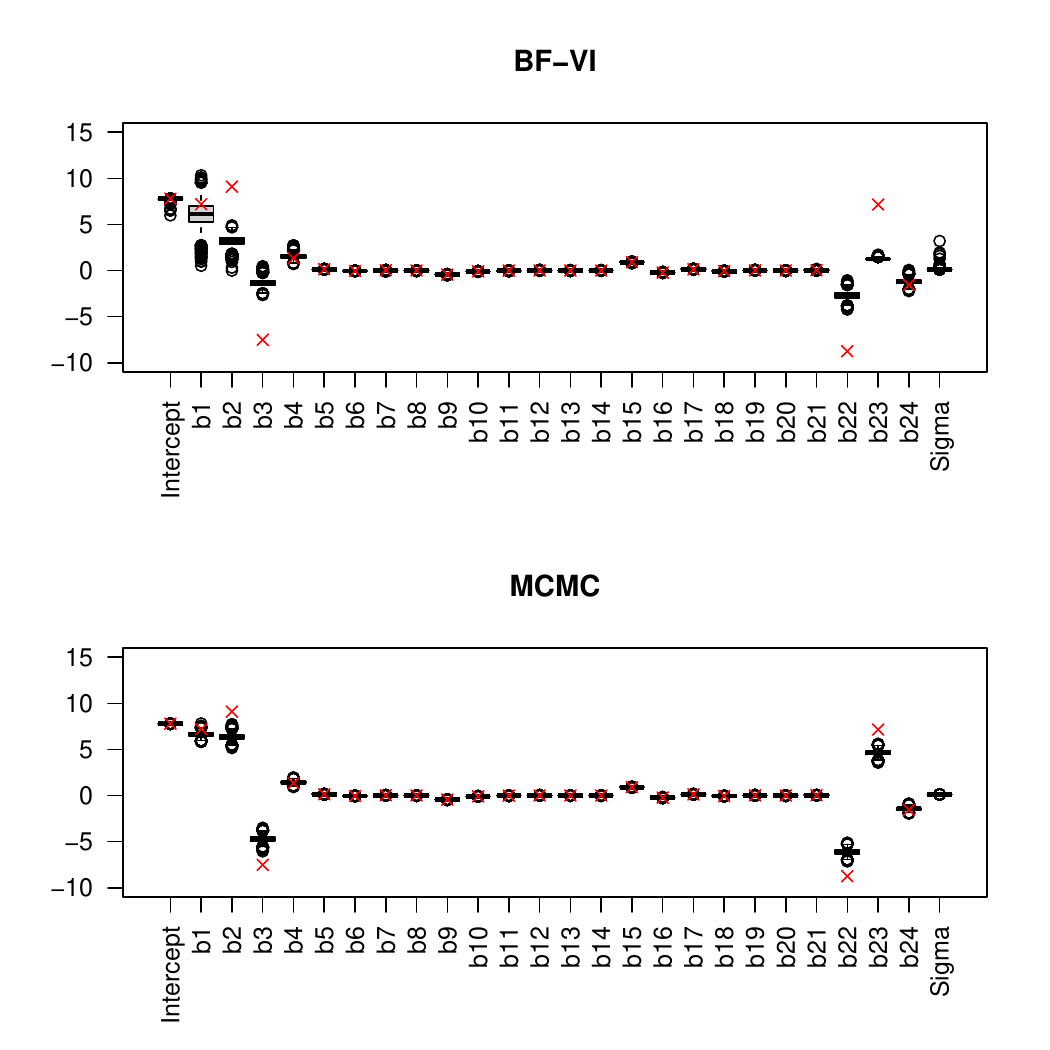}
    \caption{Comparison of samples from the marginals of MCMC samples and samples from the approximative posterior for the diamond data set. Here, the MCMC solution is very narrow, and the BF-VI (as other approximations) fails to converge to the narrow posterior. The red crosses indicate the maximum likelihood solution.}
    \label{fig:diamonds}
\end{figure}

\subsection{8School}
The 8-Schools example comes in NCP and CP parameterizations. See Table~\ref{tab:models} for the model definition, including prios.

\subsubsection{Additional Results}
In Fig.~\ref{fig:8schools}, we show samples from the marginals of the 8-dimensional parameter $\theta$ (right side, centered parameterization) and $\tilde{\theta}$ (left side, non-centered) parameterization. For the MCMC samples, the easier centered parameterization yielding $\tilde{\theta}$ is used for both cases, and $\theta$ is calculated from it. We see an underestimation of the true posterior.
\begin{figure}[h]
    \includegraphics[width=0.45\textwidth] {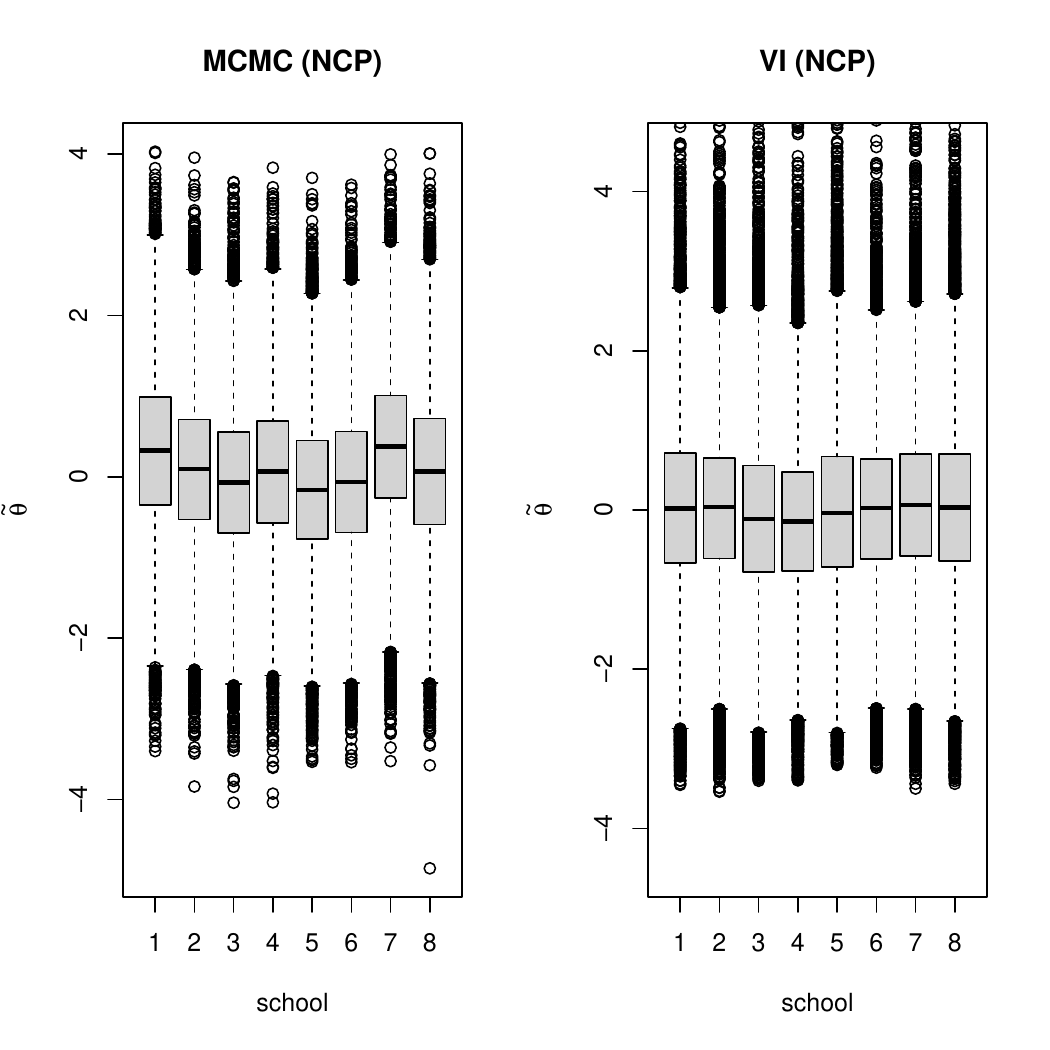}
    \includegraphics[width=0.45\textwidth] {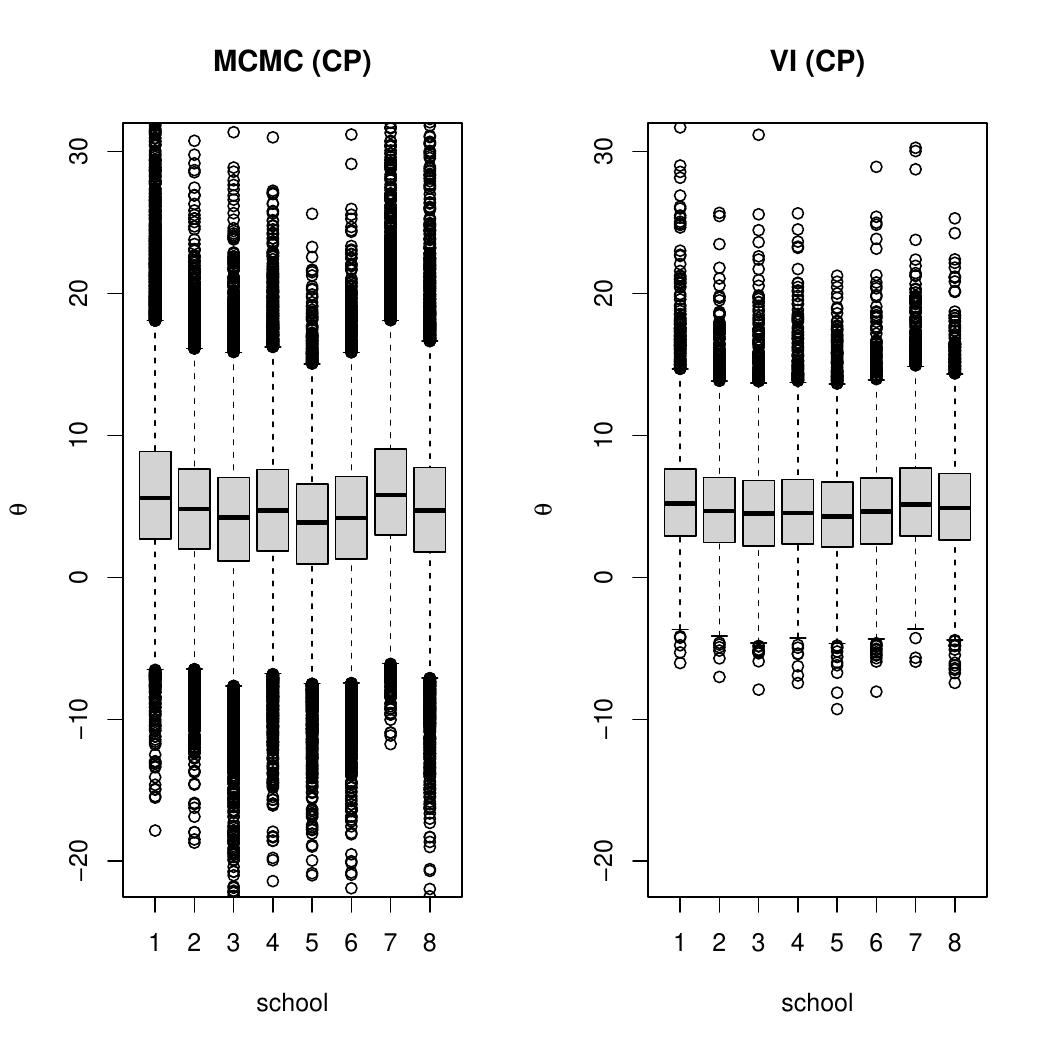}
    \caption{Marginal samples from MCMC and BF-VI for the 8 schools examples. For the NCP-parametrization $\tilde{\theta}_i$ ($i=1,\ldots,8$) is shown on left two plots and $\theta$ right two plots (CP-Parametrization). }
    \label{fig:8schools}
\end{figure}


\subsection{NN-Based Non-Linear Regression Example}
\subsubsection*{The Actual Data}
We have $N=9$ samples with\newline $x=-5.412390, -4.142973, -5.100401, -4.588446, -2.057037, -2.003591, -3.740448, \newline -5.344997,  4.506781$ and\newline $y =  0.9731220,  0.9664410,  1.2311585, 0.5193988, -1.2059958, -0.9434611, \newline  0.8041748,  0.8299642, -1.3704962
$.
\subsubsection*{Details of the Model}
The model is a simple NN with one hidden layer comprising 3
neurons and one neuron in the output layer, giving the
conditional mean. All weights have a $N(0,1)$ prior. 

\begin{figure}[h]
  \centering
  \includegraphics[width=0.8\linewidth]{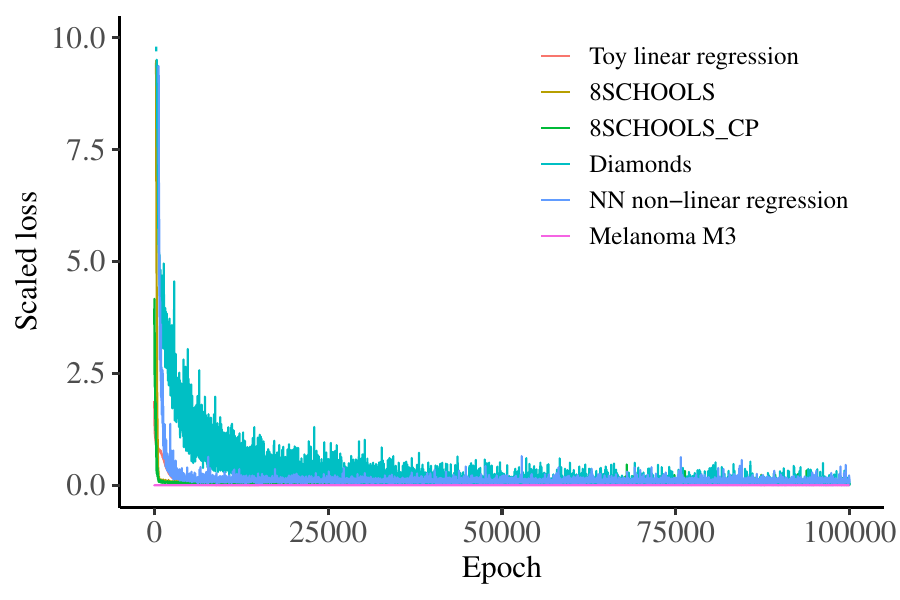}
  \caption{Loss curve for the different experiments. The loss is scaled by subtracting the minimum loss and dividing by the absolute value of the minimum loss.}
  \label{fig:loss_all}
\end{figure}

\ifstan
\subsection{Stan Code}
To unambiguously define the models, we provide the Stan code in the following.

\begin{lstlisting}[language=Stan, caption="Stan code for the Chauchy example", label={stan:chauchy}]
data{
  int<lower=0> N;
  real<lower=0> gamma;
  vector[N] y;
}
parameters{
  real xi;
}
model{
  y ~ cauchy(xi, gamma);
  xi ~ normal(0, 1);
}
\end{lstlisting}

\begin{lstlisting}[language=Stan, caption="Stan code for the toy linear regression example", label={stan:toylr}]
data {
  int<lower=0> N;
  int<lower=1> P;
  vector[N] y;
  matrix[N,P] x;
}

parameters {
  vector[P] w;
  real b;
  real<lower=0> sigma;
}

model {
  y ~ normal(x * w + b, sigma);
  b ~ normal(0,10);
  w ~ normal(0, 10);
  sigma ~ lognormal(0.5,1);
}
\end{lstlisting}

\begin{lstlisting}[language=Stan,caption="Stan code for the Diamond Example", label=lst:diamond]
// The code has been taken from https://github.com/stan-dev/posteriordb
// generated with brms 2.10.0
functions {
}
data {
  int<lower=1> N;  // number of observations
  vector[N] Y;  // response variable
  int<lower=1> K;  // number of population-level effects
  matrix[N, K] X;  // population-level design matrix
  int prior_only;  // should the likelihood be ignored?
}
transformed data {
  int Kc = K - 1;
  matrix[N, Kc] Xc;  // centered version of X without an intercept
  vector[Kc] means_X;  // column means of X before centering
  for (i in 2:K) {
    means_X[i - 1] = mean(X[, i]);
    Xc[, i - 1] = X[, i] - means_X[i - 1];
  }
}
parameters {
  vector[Kc] b;  // population-level effects
  // temporary intercept for centered predictors
  real Intercept;
  real<lower=0> sigma;  // residual SD
}
transformed parameters {
}
model {
  // priors including all constants
  target += normal_lpdf(b | 0, 1);
  target += student_t_lpdf(Intercept | 3, 8, 10);
  target += student_t_lpdf(sigma | 3, 0, 10)
    - 1 * student_t_lccdf(0 | 3, 0, 10);
  // likelihood including all constants
  if (!prior_only) {
    target += normal_id_glm_lpdf(Y | Xc, Intercept, b, sigma);
  }
}
generated quantities {
  // actual population-level intercept
  real b_Intercept = Intercept - dot_product(means_X, b);
}
\end{lstlisting}

\begin{lstlisting}[language=Stan, caption="Stan code for 8 Schools in the NCP parameterization", label=stan:NCP]
//eight_schools_ncp.stan
data {
  int<lower=0> J;
  real y[J];
  real<lower=0> sigma[J];
}

parameters {
  real mu;
  real<lower=0> tau;
  real theta_tilde[J];
}

transformed parameters {
  real theta[J];
  for (j in 1:J)
    theta[j] = mu + tau * theta_tilde[j]; //theta[j] ~ N(mu, tau*theta_tilde[j])
}

model {
  mu ~ normal(0, 5);
  tau ~ cauchy(0, 5);
  theta_tilde ~ normal(0, 1);
  y ~ normal(theta, sigma);
}
\end{lstlisting}

\begin{lstlisting}[language=Stan, caption="Stan code for the 8 schools example in the CP parametrization", label=stan:CP]
//eight_schools_cp.stan
data {
  int<lower=0> J;
  real y[J];
  real<lower=0> sigma[J];
}

parameters {
  real mu;
  real<lower=0> tau;
  real theta[J];
}

model {
  //Priors for p(mu, tau, theta)
  mu ~ normal(0, 5);
  tau ~ cauchy(0, 5);
  theta ~ normal(mu, tau);
  //Likelihood
  y ~ normal(theta, sigma);
}
\end{lstlisting}

\begin{lstlisting}[caption="Stan code for the NN based non-linear regression example", language=Stan, label=lst:stan_network]
functions {
    vector calculate_mu(matrix X, matrix bias_first_m, 
        real bias_output, matrix w_first, vector w_output, int num_layers) {
		int N = rows(X);
		int num_nodes = rows(w_first);
		matrix[N, num_nodes] layer_values[num_layers - 2];
		vector[N] mu;
        layer_values[1] = inv_logit(bias_first_m + X * w_first');   
		mu = bias_output + layer_values[num_layers - 2] * w_output;
      return mu;
    }
  }
  data {
    int<lower=0> N;						// num data
    int<lower=0> d;						// dim x
    int<lower=0> num_nodes;				// num hidden unites
    int<lower=1> num_middle_layers;		// num hidden layer
    matrix[N,d] X;						// X
    real y[N];							// y
	int<lower=0> Nt;					// num predicive data
	matrix[Nt,d] Xt;					// X predicive
	real<lower=0> sigma;				// const sigma
  }
  transformed data {
    int num_layers;
    num_layers = num_middle_layers + 2;
  }
  parameters {
    vector[num_nodes] bias_first;
    real bias_output;
    matrix[num_nodes, d] w_first;
    vector[num_nodes] w_output;
	// hyperparameters
    real<lower=0> bias_first_h;
    real<lower=0> w_first_h;
    real<lower=0> w_output_h;
  } 
  transformed parameters {
    matrix[N, num_nodes] bias_first_m = rep_matrix(bias_first', N);
  }
  model{
    vector[N] mu;
    mu = calculate_mu(X, bias_first_m, bias_output, w_first, w_output, num_layers);
    y ~ normal(mu,sigma);
    //priors
    bias_first_h ~ normal(0, 1);
    bias_first ~ normal(0, 1);
    bias_output ~ normal(0, 1);
    w_first_h ~ normal(0, 1);
    to_vector(w_first) ~ normal(0, 1);
    w_output_h ~ normal(0, 1);
    w_output ~ normal(0, 1);
  }
  generated quantities{
    vector[Nt] predictions;
	matrix[Nt, num_nodes] bias_first_mg = rep_matrix(bias_first', Nt);
	vector[Nt] mu;
	mu = calculate_mu(Xt, bias_first_mg, bias_output,w_first, w_output, num_layers);
	for(i in 1:Nt){ 
		predictions[i] = normal_rng(mu[i],sigma);
	}
   }
\end{lstlisting}
\fi
\fi
\end{document}